\documentclass[sigconf]{acmart}

\renewcommand\footnotetextcopyrightpermission[1]{} 
\usepackage{fancyhdr}
\usepackage{hyperref}
\usepackage{cleveref}
\AtBeginDocument{%
  }

\begin{document}

\title{User-Friendly Customized Generation with Multi-Modal Prompts}

\author{Linhao Zhong}
\email{zhongzero@sjtu.edu.cn}
\affiliation{%
  \institution{Shanghai Jiao Tong University}
  \city{Shanghai}
  \country{China}
}

\author{Yan Hong}
\email{yanhong.sjtu@gmail.com}
\affiliation{%
  \institution{AntGroup}
  \city{Hangzhou}
  \country{China}
}

\author{Wentao Chen}
\email{leonard_chen@sjtu.edu.cn}
\affiliation{%
  \institution{Shanghai Jiao Tong University}
  \city{Shanghai}
  \country{China}
}

\author{Binglin Zhou}
\email{zhoubinglin@sjtu.edu.cn}
\affiliation{%
  \institution{Shanghai Jiao Tong University}
  \city{Shanghai}
  \country{China}
}

\author{Yiyi Zhang}
\email{yi95yi@sjtu.edu.cn}
\affiliation{%
  \institution{Cornell University}
  \city{New York}
  \country{America}
}

\author{Jianfu Zhang}
\email{c.sis@sjtu.edu.cn}
\affiliation{%
  \institution{Shanghai Jiao Tong University}
  \city{Shanghai}
  \country{China}
}

\author{Liqing Zhang}
\email{zhang-lq@cs.sjtu.edu.cn}
\affiliation{%
  \institution{Shanghai Jiao Tong University}
  \city{Shanghai}
  \country{China}
}


\begin{abstract}
  Text-to-image generation models have seen considerable advancement, catering to the increasing interest in personalized image creation. Current customization techniques often necessitate users to provide multiple images (typically 3-5) for each customized object, along with the classification of these objects and descriptive textual prompts for scenes. This paper questions whether the process can be made more user-friendly and the customization more intricate. We propose a method where users need only provide images along with text for each customization topic, and necessitates only a single image per visual concept. We introduce the concept of a ``multi-modal prompt'', a novel integration of text and images tailored to each customization concept, which simplifies user interaction and facilitates precise customization of both objects and scenes. Our proposed paradigm for customized text-to-image generation surpasses existing finetune-based methods in user-friendliness and the ability to customize complex objects with user-friendly inputs. Our code is available at \href{https://github.com/zhongzero/Multi-Modal-Prompt}{https://github.com/zhongzero/Multi-Modal-Prompt}.
\end{abstract}

\maketitle

\section{Introduction}

\begin{figure}[ht]
  \centering
  \includegraphics[width=\linewidth]{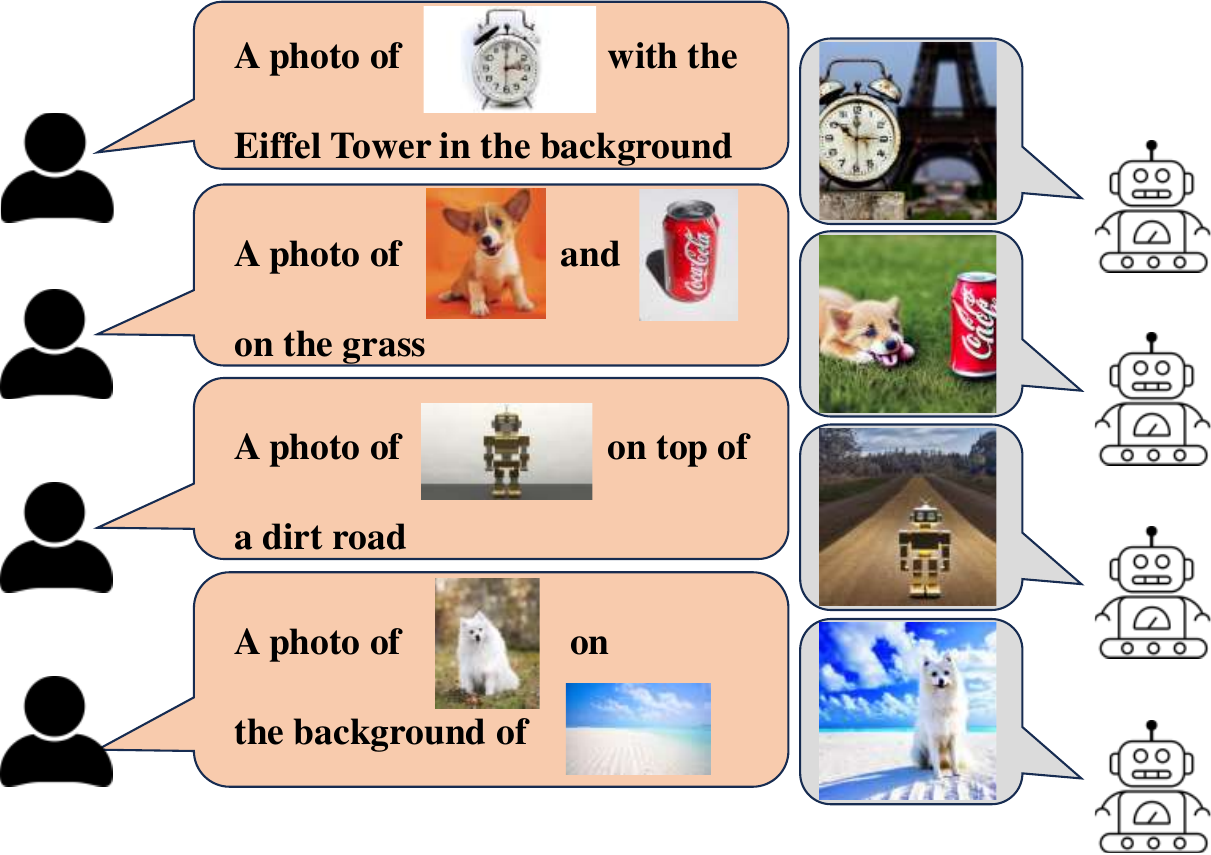}
  \caption{\textbf{User-Friendly Customization Through Multi-Modal Prompts:} Leveraging multi-modal prompts enables users to precisely tailor both objects and scenes of interest. When provided with such prompts, our paradigm efficiently generates images that not only feature the specified objects within the desired scenes but also excel in the detailed customization of complex objects, showcasing our method's superior performance and user-centric approach.} 
  \label{fig:introduction_overview}
\end{figure}


As the field of text-to-image generation advances, there's an increasing demand for the capabilities of customized image generation. To date, several methodologies \cite{textual-inversion, dreambooth, custom-diffusion, blip_diffusion, subject_driven_generation} have been proposed to create images from textual descriptions as well as from specific images. Among these, Textual Inversion \cite{textual-inversion} stands out as a seminal approach for customized text-to-image generation. This method enables users to customize the output by a specific object and textual prompt. More precisely, by using a small set of images depicting the same object, Textual Inversion leverages the finetuning process of diffusion models to convert these images into a unique token. This token can then be used to consistently generate images of the object.
Following in the footsteps of Textual Inversion, Dreambooth \cite{dreambooth} and Custom Diffusion \cite{custom-diffusion} introduce further refinements for more detailed customization. They extend the basic method by asking for the specification of an object's class and utilizing this class name to maintain the object's class-specific prior knowledge throughout the finetuning process. These approaches are all rooted in finetune-based methodologies, necessitating a finetuning phase to translate the knowledge from provided images into textual tokens. They require a small collection of images depicting the same object—referred to as ``few-shot'' learning here—to acquire a comprehensive understanding of the object. Moreover, the specification of the object's class is essential, with even Textual Inversion necessitating the class of the object as an initial condition for the finetuning process to enhance the outcome.


Current finetune-based methods for customized text-to-image generation, while innovative, fall short in terms of user-friendliness and the granularity of customization. A primary issue lies in the requirement for multiple images; many existing approaches \cite{textual-inversion,dreambooth,custom-diffusion} necessitate between three to five images, or even more, for effective customization. 
Often, users possess only a single image of the object they want to customize, a scenario we refer to as ``one shot''. In such cases, the output quality of these finetune-based methods tends to be insufficient.
Furthermore, the obligation to specify the class of the object introduces an additional barrier to accessibility for users. For certain objects, determining and defining their precise class can be a challenge for users. Besides, while current methods may be able to customize relatively simple objects, they lack the capability needed for more complex objects, resulting in customization that may not fully capture the intricate details desired by users.


In this study, we introduce an innovative paradigm for customized text-to-image generation that builds upon existing finetune-based methods. Our approach emphasizes a ``one-shot'' customization process and does not require users to specify the objects' class. This advancement significantly simplifies user interaction for generating customized images, necessitating only a single multi-modal prompt from users. We define ``multi-modal prompt'' here as a prompt that incorporates both textual and visual elements. This concept is illustrated in \cref{fig:introduction_overview}, where users can incorporate images containing the objects they wish to customize directly into the prompt, alongside conventional text syntax to delineate the desired scene for customization.
Our paradigm offers a more general approach to customization compared to existing finetune-based methods, without the need for additional information about the object's class. Furthermore, the enhancement in the customization of complex objects is particularly notable, where our method demonstrates superior capability in capturing and reflecting intricate details. By simplifying the input requirements to a single, integrated prompt that seamlessly blends text and imagery, we make the process of generating tailored images more accessible and user-friendly, opening up new possibilities for customized image generation.

\section{Related Works}

\subsection{Text-to-Image generation}


Text-to-image generation is a multi-modal task at the intersection of computer vision and natural language processing, aiming to create images from textual descriptions. Contemporary methods primarily fall into two categories: GAN-based \cite{GAN} and diffusion-based \cite{DDPM, IDDPM, DDIM, diffusion_survey}. GAN-based approaches, such as Text-Conditional GAN \cite{Text-Conditional-GAN}, StackGAN \cite{StackGAN} and AttnGAN \cite{AttnGAN}, leverage generative adversarial networks to produce images from text. Conversely, diffusion-based methods, exemplified by Stable Diffusion \cite{SD}, DALL-E 2 \cite{DALLE2}, and Imagen \cite{Imagen}, utilize diffusion models for image generation. A thorough review of these methodologies is provided in \cite{text-to-image-survey}. Currently, many novel methods \cite{attend_and_excite, ediffi, versatile_diffusion, raphael, cross_attention_guidance, muse, spatext, vector_quantized_diffusion, de_fake} have been proposed to improve the performance of text-to-image generation and explore solutions to the problems of current methods. A closely related research area is image editing \cite{instructpix2pix, imagic, editing_inversion, paint_by_example}, where the goal is to modify an image based on textual descriptions or other conditioning inputs. Also, current research \cite{controlnet, controllable_generation, uni-controlnet} explores image generation with more general conditioning inputs. There are many applications based on text-to-image generation, including video generation \cite{tune_a_video, video_synthesis, video_preserve, stablevideo, align_your_latents}, 3D object generation \cite{magic3d, mvdream, dream3d, rodin, dreameditor} and others \cite{image_to_image, voynov2023sketch}. Text-to-video generation and text-to-3D generation are also two important research areas with more complex and challenging tasks, and much of the research in these areas is based on text-to-image generation, which is a relatively mature research area.

\subsection{Customized Text-to-Image generation}


Advancements in text-to-image generation have sparked interest in customization, where the goal is to generate varied images of the same object across different scenes, based on one or a few reference images. This area has witnessed the development of finetune-based approaches, and many current methods are characterized by a finetuning process that translates knowledge from given images into text tokens for subsequent image generation. Key methodologies include Textual Inversion \cite{textual-inversion}, which learns a unique token embedding for each object; Dreambooth \cite{dreambooth}, which finetunes a model's entire backbone to assimilate given image knowledge while preserving objects' class prior knowledge; and Custom Diffusion \cite{custom-diffusion}, an approach that merges the similar strategies of Textual Inversion and Dreambooth for enhanced customization. Additionally, \cite{blip_diffusion, subject_driven_generation, domain-agnostic, mix_of_show, continual_diffusion, p+, disenbooth, cones, key_locked_rank, break_a_scene} explore novel methods for the task of customized text-to-image generation. To address the problem of slow generation speed and high memory consumption, existing works \cite{fast_personalization, vico, hyperdreambooth, instantbooth, elite} focus on faster generation with less memory consumption for customized text-to-image generation. There are applications based on customized text-to-image generation, including concept ablation \cite{ablation_concept} and animation generation \cite{animatediff}. 

\subsection{Image Captioning}
\label{sec:image-captioning}


Image captioning, a multi-modal task akin to text-to-image generation, is essentially an image-to-text conversion process. It involves generating appropriate textual descriptions, or captions, for given images. Early image captioning efforts included template-based methods, which utilized templates with blanks to be filled in based on image content, as seen in works like \cite{image-captioning-template1, image-captioning-template2}. Retrieval-based methods. On the other hand, retrieval-based methods identified captions by finding similar images to the given one and borrowing their captions, exemplified by \cite{image-captioning-retrieval1, image-captioning-retrieval2, image-captioning-retrieval3}. Presently, the dominant approaches in image captioning employ neural networks, with a particular emphasis on leveraging pre-trained models from vision-language tasks. BLIP \cite{BLIP} stands out as a notable example of using pre-trained models to accurately generate captions for images.

\subsection{Large Language Model}
\label{sec:LLM}


Large language models (LLMs) are developed through training on extensive text corpora and are instrumental in a variety of natural language processing (NLP) tasks. Prominent examples include GPT-3 \cite{GPT3}, InstructGPT \cite{InstructGPT}, BERT \cite{BERT}, Llama 2 \cite{llama2} and T5 \cite{T5}. Among these, ChatGPT, built upon InstructGPT, has garnered attention for its proficiency in generating human-like dialogues, making it a valuable tool for dialogue generation and other NLP applications. ChatGPT's effectiveness stems from its extensive knowledge, acquired through learning from large-scale text data. Several works \cite{misinformation, misinformation2, llm_security_survey, judging_llm} are exploring and discussing the capabilities and limitations of large language models. Also, multi-modal LLM \cite{multimodal_llm, multimodal_llm2, X_llm, shikra} has attracted more and more attention, which is a combination of LLM and other modalities, such as image, audio, video, etc. 
\section{Methodology}

In this section, we introduce a novel paradigm for customized text-to-image generation that refines current finetuning methodologies to enhance user interaction through the use of multi-modal prompt. 
Our paradigm is designed to facilitate user-friendly image customization, requiring users to provide only a single specific image for each concept alongside textual prompts. 
This contrasts with previous approaches that often necessitate multiple images (typically 3-5) to adequately capture one concept. Our approach acknowledges that users may not always have access to multiple images of the same concept. Thus, a single image input for each concept represents a significant advancement in making image customization more accessible. 
Furthermore, our approach does not require additional information like objects' class names. 
By leveraging multi-modal prompts that combine this single image with text, we aim to accurately capture and generate images reflecting the specified concept, even if the input image is complex. 
The methodology and its implementation are detailed in the subsequent subsections.

\subsection{Multi-Modal Prompt}
\begin{figure}[ht]
  \centering
  \includegraphics[width=8cm]{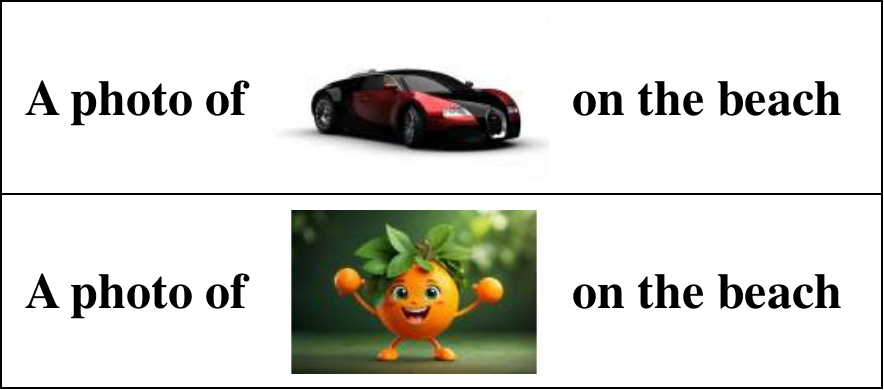}
  \caption{\textbf{Two Examples of Multi-Modal Prompts:} The first example features an image of a car showcasing a red and black color scheme, isolated against a void of other objects or background elements. The second example displays a cartoon orange positioned on a road, set against a forest backdrop, illustrating the versatility of multi-modal prompts in depicting complex and diverse scenarios.}
  \label{multi-modal-prompt-example}
\end{figure}

It is crucial to articulate a precise definition of what constitutes a multi-modal prompt, a concept central to our paradigm. \cref{multi-modal-prompt-example} showcases two illustrative examples of multi-modal prompts, facilitating a deeper understanding of their structure and intended user interaction.
The first example features an image of a car with a red and black color scheme, isolated from any background elements. This simplicity allows for a straightforward interpretation: the user's intent is to generate an image of this specific car situated in a novel scene. 
The second example presents a cartoon orange positioned on a road with a forest backdrop. While the user's aim is to similarly transplant the main subject (the cartoon orange) into a new scene, 
the inclusion of a complex background poses a challenge for straightforward customization.
Such scenarios, featuring complex backgrounds, are commonplace. Recognizing the potential difficulty for users in background removal, our paradigm is designed to obviate this necessity, enhancing user accessibility.
Further examination of datasets from leading customized text-to-image generation research—such as Textual Inversion \cite{textual-inversion}, Dreambooth \cite{dreambooth} and Custom Diffusion \cite{custom-diffusion}—reveals a common user inclination towards customizing primary objects within images. 
This observation underpins our assumption that users primarily seek to customize the main subjects of their images.
In this work, our paradigm interprets multi-modal prompts through the lens of pure text prompts. These are derived by transforming visual depictions of main subjects into textual descriptions, which then constitute the backbone of the customization process. Through this methodology, we facilitate a user-friendly and intuitive interface for personalized image generation, significantly simplifying the customization process.

\begin{figure*}[t]
  \centering
  \includegraphics[width=0.95\linewidth]{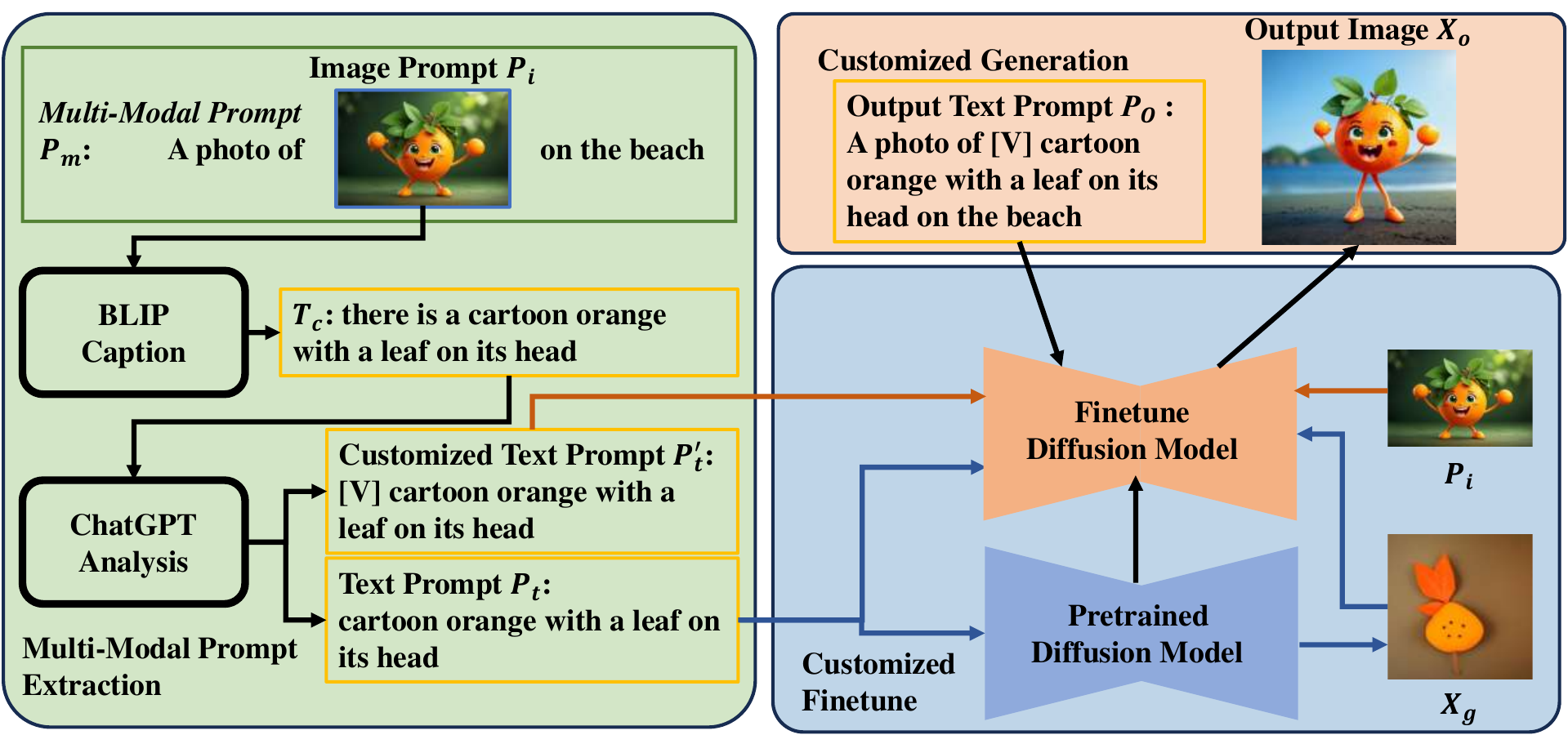}
  \caption{\textbf{The overview of our paradigm.} Our innovative paradigm is divided into two crucial components: the extraction of main object descriptions and the customization of concepts while preserving detailed prior knowledge. Initially, the process involves extracting descriptions of the main objects within the multi-modal prompt images, which is executed in two phases: image captioning using BLIP, followed by semantic analysis with ChatGPT. Subsequently, the second component utilizes these extracted descriptions to maintain the detailed prior knowledge of the main objects, thereby enhancing the customization performance.}
  \label{method_overview}
\end{figure*}

Our paradigm is structured around two core components: extraction of main object descriptions and customization of concepts with the retention of detailed prior knowledge. These components will be elaborated upon in the subsequent subsections. 
\cref{method_overview} illustrates an overview of our approach, demonstrating the process involving an image prompt $P_i$ within the multi-modal prompt $P_m$. 
This example serves as a foundational scenario to facilitate comprehension of our paradigm. The methodology for handling multi-modal prompt $P_m$ containing multiple images shares similarities with this foundational scenario, which will be discussed in further detail at a later stage.

\subsection{Multi-Modal Prompt Extraction}

The first part of our paradigm entails the extraction of the text prompt describing the main object from the image prompt $P_i$ in the multi-modal prompt $P_m$, denoted as $P_t$. 
This is accomplished through a meticulously designed two-stage process. Initially, for the image $P_i$ within the multi-modal prompt, we undertake the task of generating an accurate text description of the entire image, a process known as image captioning. This comprehensive text description is identified as $T_c$. 
Subsequently, the syntax of the text description $T_c$ is analyzed to obtain the description of the main object $P_t$ present in the image, a phase termed semantic analysis.



\subsubsection{Image Captioning}

As discussed in \cref{sec:image-captioning}, BLIP is a representative work that uses pre-trained models for the image captioning task. Its proficiency in generating precise captions for given images makes it an ideal choice for our purposes. Utilizing BLIP, we generate the text description $T_c$ for the image $P_i$ within the multi-modal prompt $P_m$. 
For instance, as illustrated in \cref{method_overview}, the multi-modal prompt at the top of the figure undergoes image captioning, resulting in an accurate text description $T_c$: ``there is a cartoon orange
with a leaf on its head'' for the image. This example underscores BLIP's capability to produce detailed and relevant captions that are instrumental in further processing steps.

\subsubsection{Semantic Analysis}

Upon acquiring the image's text description $T_c$, the subsequent task involves conducting semantic analysis to obtain the description of the main object. Traditional methods struggle with this task due to the lack of a fixed format in the text descriptions generated by BLIP. 
In light of this challenge, we pivot towards utilizing large language models for semantic analysis. As elaborated in \cref{sec:LLM}, large language models are pre-trained on extensive text corpora, enabling them to grasp the semantic content of text descriptions proficiently. 
For this purpose, we employ ChatGPT, a leading example of such models. The process begins by introducing ChatGPT to the specific nature of our semantic analysis task through examples, thus training it to understand and execute the task. 
Our objective is for ChatGPT to discern and extract elements such as the foreground (main object's description $P_t$), background, and actions from $T_c$, with the understanding that background and action may not always be present. 
This approach, focusing on extracting a comprehensive set of elements rather than solely the main object, enhances ChatGPT's learning process and improves the accuracy of the results. 
The detailed version of the prompts provided to ChatGPT are included in the supplementary material. 


  
  

  
  


It should also be noted that the responses from ChatGPT can vary in format, even when identical prompts are provided. To address this variability, we implement a multiple-inquiry mechanism. 
This approach involves posing the same question to ChatGPT several times, filtering out responses in unexpected formats, and selecting the most frequently occurring answer as the definitive result. 
This mechanism has demonstrated efficacy in our experiments. 



\subsection{Concept Customization with Multi-Modal Prompt}

In this part, we introduce the process of concept customization with a focus on preserving detailed prior knowledge. Our approach leverages diffusion models to internalize the concept of the main object depicted in the image within the multi-modal prompt. This internalization is facilitated through a finetuning process, where diffusion models are adjusted to enhance their ability to generate customized images based on output text prompts $P_o$. 
Central to our method is the principle of detailed prior knowledge preservation, a strategy inspired by methods like Dreambooth \cite{dreambooth} and Custom Diffusion \cite{custom-diffusion}. 
These techniques ensure that the finetuned model retains a deep understanding of the original concept while being adept at applying this knowledge to generate images that align closely with the user's specifications.

\subsubsection{Preliminary: Diffusion Models}

Diffusion models \cite{DDPM, IDDPM, DDIM} are a type of generative models capable of transforming noise into a coherent sample that mirrors a target distribution through a sequence of denoising steps.
These models are also adept at conditional generation, where the generation process is guided by a specified condition. 
Let $q(x)$ be the data distribution, with $x$ being a sample from $q(x)$ and $c$ representing the condition. 
$x_t:=\alpha_t x + \sigma_t \epsilon$ is obtained by adding noise to $x$ for $t$ steps, where $\epsilon$ is a noise sample from the Gaussian distribution, $t \in \left\{ 1,2,\ldots,T \right\}$ is the timestep, and $\alpha_t$ and $\sigma_t$ are the hyper-parameters of the diffusion model. 
Let $\epsilon_{\theta}$ be the model. The training objective of the diffusion model is to minimize mean squared error, formalized as:

\begin{equation}
  \min_{\theta} \mathbb{E}_{x, c, t, \epsilon} \left[ \left\| \epsilon - \epsilon_{\theta}(x_t, t, c) \right\|_2^2 \right].
\end{equation}


Within text-to-image diffusion models, the data distribution $q(x)$ specifically pertains to the distribution of images, while the condition $c$ is defined by the text prompt. During the training phase, the image $x$ and the text prompt $c$ are provided as paired entities. 
This pairing facilitates the model's learning process, enabling it to generate images that are closely aligned with the semantic content of the given text prompt. By conditioning the generation process on text prompts, the model leverages linguistic cues to accurately render visual representations, embodying the described scenarios or objects.

\subsubsection{Finetuning Process}


The finetuning process for a text-to-image diffusion model is crucial for adapting the model to specific tasks. This process involves encoding the knowledge contained within given images into textual representations, utilizing pairs of image $x$ and text prompt $c$. 
The objective during finetuning is to adjust the model so that it can more accurately generate images corresponding to new text prompts based on previously learned image-text associations. 
The loss function used in the finetuning process is designed to minimize the discrepancy between the original noise $\epsilon$ and the model-generated noise $\epsilon_{\theta}$, conditioned on the image $x$, timestep $t$, and condition $c$. It is represented as follows:
\begin{equation}
  \mathcal{L}_{x,c} = \mathbb{E}_{t, \epsilon} \left[ \left\| \epsilon - \epsilon_{\theta}(x_t, t, c) \right\|_2^2 \right].
\end{equation}
This loss function serves as the guiding metric for the finetuning, emphasizing the model's ability to accurately interpret and render images from textual descriptions by closely aligning the model's output with the semantic intent of the text prompts.


For effective customized text-to-image generation, finetune-based methods must adeptly incorporate the knowledge of specific images. These methods often employ a unique token, such as $[V]$, as a placeholder for the image-specific knowledge within the condition $c$, pairing it with the image $x$ for the finetuning of the diffusion model. However, this approach presents limitations, primarily because it struggles to precisely associate the knowledge of the main object in the given image $P_i$ with the special token. Instead, it may inadvertently merge the broader knowledge of the entire image $P_i$ with the token. 
This misalignment can cause generated images to either overfit to the entirety of the provided image or fail to accurately and meticulously depict the customized objects. Such outcomes underscore the necessity for a more sophisticated method that can ensure precise and detailed object representation in the generated images.
In response to this, we propose a paradigm adaptable to finetune-based customized image generation methods like Dreambooth \cite{dreambooth} and Custom Diffusion \cite{custom-diffusion}, focusing here on its application within the Dreambooth framework due to space constraints. For details on its application to Custom Diffusion, please see the supplementary material. 
Our method employs a parallel strategy to conserve the intricate prior knowledge of customized objects, thereby addressing the aforementioned limitations. 
By leveraging the precise and detailed description $P_t$ of the customized object, we can better preserve prior knowledge, thereby enhancing customization performance. Specifically, we introduce a composite descriptor $P_t' = \text{``} [V] \quad P_t \text{''}$ that incorporates a special token alongside $P_t$ to uniquely identify the customized object. In this setup, $P_i$ represents the given image, and $X_g$ is the image generated by the pretrained diffusion model using the text prompt $P_t$. A hyper-parameter $\lambda$ is employed to balance the components of the loss function during finetuning:

\begin{equation}
  \mathcal{L} = \mathbb{E}_{X_g} \left[ \mathcal{L}_{P_i,P'_t} + \lambda \mathcal{L}_{X_g,P_t} \right].
\end{equation}


This approach utilizes the accurate and detailed description of the customized object to impart a deeper understanding of the object's unique attributes to the diffusion model. During finetuning, the model focuses on assimilating the distinctive characteristics of the customized object beyond what is typically generated using $P_t$ on a pretrained model. The inclusion of a specific and detailed description allows the special token to more effectively highlight the unique aspects of the customized object, resulting in more intricately detailed representations, particularly for complex objects.

\subsubsection{Customized Generation}


After the finetuning process, our finetuned diffusion model is now adept at generating images based on the knowledge of the customized object. To facilitate customized image generation, we leverage a pure text prompt $P_o$. 
This is accomplished by substituting the image $P_i$ in the multi-modal prompt $P_m$ with the composite descriptor $P_t'$, thereby deriving $P_o$. 
The descriptor $P_t'$ encapsulates the essential knowledge of the customized object. By inputting $P_o$ into our finetuned diffusion model, we direct the model to generate images that not only feature the customized object but also integrate it within the specified scene, achieving a harmonious blend of customization and context.

\subsection{Multiple Images Situation}

The situation of handling a multi-modal prompt $P_m$ that contains multiple images is similar to the process for a single image, with tailored adjustments to accommodate the increased complexity. 
Here, we provide a more formal explanation of the multiple images situation. Suppose the multi-modal prompt $P_m$ comprises $n$ images $P_{i_1}, P_{i_2}, ..., P_{i_n}$. In the initial stage, dedicated to main object description extraction, we employ the same process outlined in the Methodology section (Sec. 3) for each image $P_{i_j}$ within the multi-modal prompt. This process generates a text description $T_{c_j}$ for image $P_{i_j}$, from which we obtain the description $P_{t_j}$ of the main object. Consequently, we derive the descriptions $P_{t_1}, P_{t_2}, ..., P_{t_n}$ representing the main objects in all images.

Moving on to the second stage, where customization of concepts with the retention of detailed prior knowledge takes place, for each image $P_{i_j}$ and its corresponding main object's extracted description $P_{t_j}$, we introduce $P'_{t_j}= \text{``} [V] \quad P_{t_j} \text{''}$. Additionally, let $X_{g_j}$ represent the image generated by the pretrained diffusion model using the text prompt $P_{t_j}$. Subsequently, we follow the same process detailed in the Methodology section to finetune the diffusion model using these components. The loss function for the finetuning process can be expressed as follows:

\begin{equation}
  \mathcal{L} = \sum_{j=1}^{n} \mathbb{E}_{x_{g_j}} \left[ \mathcal{L}_{P_{i_j},P'_{t_j}} + \lambda \mathcal{L}_{x_{g_j},P_{t_j}} \right]
\end{equation}

After the finetuning process, we acquire a refined diffusion model. We replace all the images $P_{i_1}, P_{i_2}, ..., P_{i_n}$ in the multi-modal prompt $P_m$ with $P'_{t_1}, P'_{t_2}, ..., P'_{t_n}$, resulting in the creation of the pure text prompt $P_o$. This text prompt, denoted as $P_o$, is then straightforwardly input into the finetuned diffusion model to generate the customized image.
\section{Experiments}

\subsection{Experimental Setup}

\subsubsection{Implementation Details}

Our paradigm is a universal paradigm for finetune-based customized text-to-image generation methods, and current mainstream finetune-based methods, such as Dreambooth \cite{dreambooth} and Custom Diffusion \cite{custom-diffusion}, can all be applied in our paradigm as the specific implementation of the part of customization of concepts with the retention of detailed prior knowledge. 
To demonstrate the effectiveness of our paradigm, we integrate Dreambooth and Custom Diffusion into our framework as Ours-Dreambooth and Ours-Custom Diffusion, respectively, for preserving detailed prior knowledge. We then benchmark their performance against the original implementations. 
We adapt Custom Diffusion by substituting its original class image retrieval approach with a generation method to better align with our paradigm. All experiments utilize the publicly available Stable Diffusion version 1.4 \cite{SD}. 
The learning rates are set to 2e-6 for both Dreambooth and Ours-Dreambooth, and 1e-5 for Custom Diffusion and Ours-Custom Diffusion. Image generation from the finetuned models is conducted with 200 inference steps and a guidance scale of 7.5.

\subsubsection{Dataset}


To demonstrate the proficiency of our paradigm in achieving detailed customization, especially with complex objects, we utilized a dataset comprising 15 diverse objects. These objects include a backpack, car, flower, and cartoon orange, representing a broad spectrum of categories. Following the approach of Dreambooth, we employed 20 recontextualization prompts for each object. 
Each prompt was associated with a single image of the object to maintain focus and consistency. For each object and prompt combination, we generated 10 images, resulting in a total of 3000 images for the experiment. This dataset and methodology proved robust for showcasing the enhanced capability of our paradigm for detailed customization.

\subsubsection{Evaluation Metrics}

We employ DINO \cite{dino} score, CLIP-I score, and CLIP-T score. These metrics collectively assess the alignment of generated images with both their corresponding images and text prompts. The DINO score calculates the average cosine similarity between the features of generated and given images within the ViT-S/16 DINO feature space, serving as a measure of image alignment. Similarly, the CLIP-I score evaluates image alignment by computing the average cosine similarity between the features of generated and given images within the CLIP \cite{clip} image feature space. For text-alignment evaluation, the CLIP-T score measures the average cosine similarity between the features of the pure text prompt and the generated images in the CLIP text/image feature space. The pure text prompt is derived by substituting the image in the multi-modal prompt with the extracted descriptions of the main objects. A detailed analysis of the accuracy of the main object's extracted description and the rationale behind the pure text prompt for text-alignment evaluation will be provided in the supplementary material.

\subsection{Comparisons with State-of-the-Art Methods}
\label{sec:comparisons-current-methods}

Here, we present the results comparing the performance of our paradigm with that of current finetune-based methods. As mentioned in the implementation details, we use the same implementation of Dreambooth and Custom Diffusion, respectively, in the part of customization of concepts with the retention of detailed prior knowledge, and they are referred to as Ours-Dreambooth and Ours-Custom Diffusion. For the original Dreambooth and Custom Diffusion, we specifically specify the class of the object.
We will show the results of the comparisons in the following subsections.

\subsubsection{Qualitative Evaluation}

\begin{figure}[tp]
  \centering
  \includegraphics[width=\linewidth]{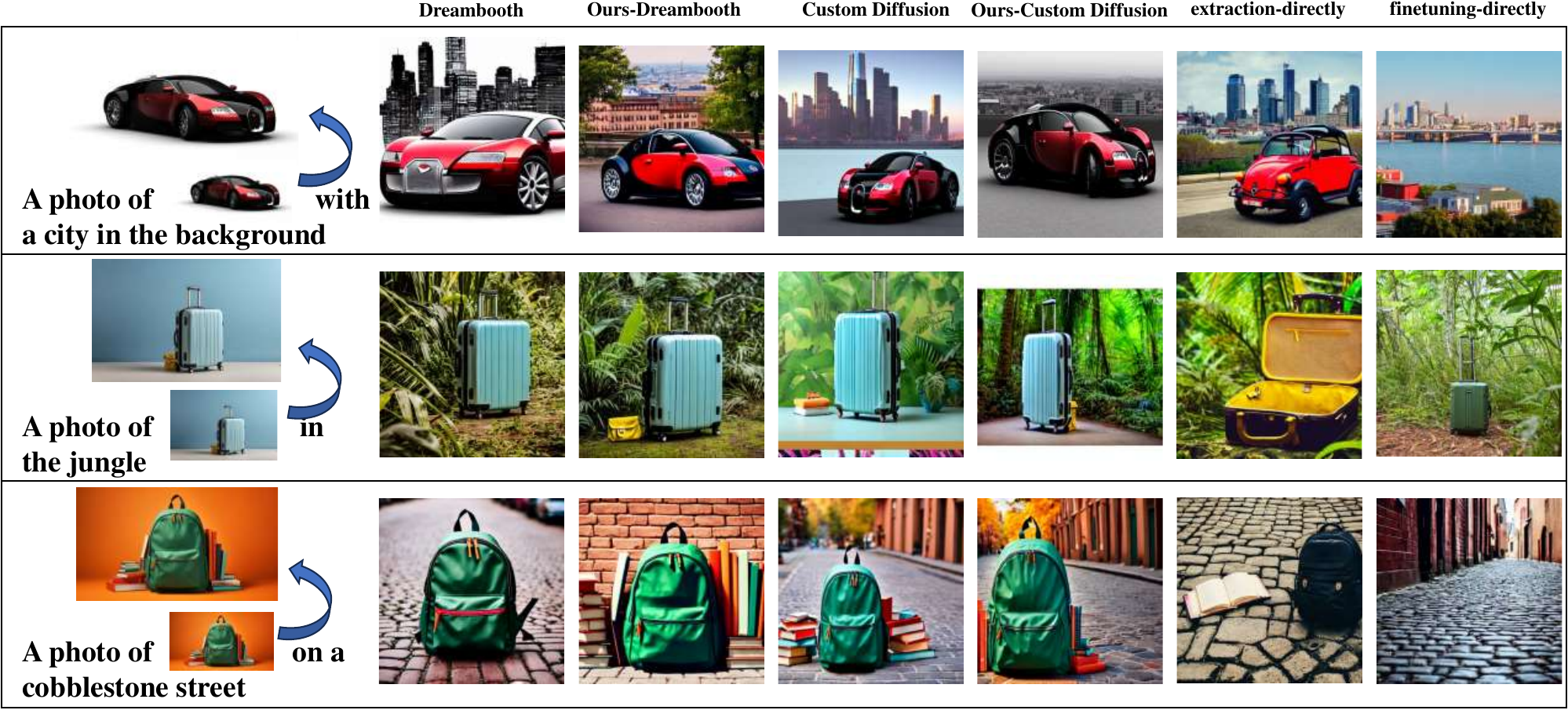}
  \caption{\textbf{Qualitative comparisons.} This figure showcases sample images generated by Dreambooth, Ours-Dreambooth, Custom Diffusion, Ours-Custom Diffusion, the extraction-directly method, and the finetuning-directly method across distinct multi-modal prompts. For an in-depth analysis, please see \cref{sec:comparisons-current-methods} and \cref{sec:ablation-study}.
  }
  \label{fig_expr}
\end{figure}

To assess the effectiveness of our methods compared to the original approaches, our focus is on customizing complex objects. The images generated from Dreambooth, Ours-Dreambooth, Custom Diffusion, and Ours-Custom Diffusion are displayed in \cref{fig_expr}. Our paradigm demonstrates superior performance in the detailed customization of complex objects, as evidenced by improved image alignment between the generated and provided images. Specifically: First Row: Dreambooth struggles to customize the car's color accurately, whereas our methods achieve precise color customization. Second Row: Dreambooth overlooks the small yellow bag adjacent to the blue suitcase. Custom Diffusion recognizes the yellow bag but does not customize it accurately. In contrast, our methods accurately customize both the small yellow bag and the blue suitcase. Third Row: Dreambooth fails to account for the books next to the backpack. Our methods, however, naturally and accurately customize the cartoon orange. These examples highlight the capacity of our methods to capture and reflect the property of complex objects more faithfully than existing finetune-based approaches.

\subsubsection{Quantitative Evaluation}


In our comprehensive quantitative assessment, we tested the efficacy of our approach versus traditional methods over 15 objects, using 20 recontextualization prompts per object. For each multi-modal prompt, we produced 10 images, culminating in an extensive dataset for our analysis.
The findings, as outlined in \cref{table_expr}, demonstrate that our paradigm consistently surpasses traditional methods across all evaluated metrics: DINO score, CLIP-I score, and CLIP-T score. These results highlight the superior capability of our approach in both image and text alignment.
Noteworthy is the remarkable improvement observed in the DINO score and CLIP-I score, signifying our method's enhanced precision in the detailed customization of complex objects. This significant progress marks a distinguished advancement in the domain of text-to-image generation, showcasing the strength of our methodology in rendering intricately detailed customizations.

\begin{table}[tp]
  \centering
  \resizebox{1\columnwidth}{!} {
  \begin{tabular}{c|c|c|c}
    \toprule
    Method & DINO score & CLIP-I score & CLIP-T score \\
    \midrule
    Dreambooth & 0.5484 & 0.8036 & 0.3120 \\
    Ours-Dreambooth & \textbf{0.6402} & \textbf{0.8743} & \textbf{0.3199} \\
    \midrule
    Custom Diffusion & 0.5714 & 0.7873 & 0.3089 \\
    Ours-Custom Diffusion & \textbf{0.7558} & \textbf{0.8808} & \textbf{0.3132} \\
    \midrule
    extraction-directly & 0.3821 & 0.7666 & \textbf{\textcolor{red}{0.3415}} \\
    finetuning-directly & 0.1053 & 0.5598 & 0.2484 \\
    \bottomrule
  \end{tabular}
  }
  \caption{\textbf{Quantitative comparisons.} This table demonstrate the performance metrics including DINO score, CLIP-I score, and CLIP-T score for Dreambooth, Ours-Dreambooth, Custom Diffusion, Ours-Custom Diffusion, the extraction-directly method, and the finetuning-directly method. For an in-depth analysis, please see \cref{sec:comparisons-current-methods} and \cref{sec:ablation-study}.
  }
  \label{table_expr}
\end{table}

\subsubsection{Human Preference Study}


To further substantiate the efficacy of our paradigm, a human preference study was conducted, comparing our methods against traditional approaches. For this study, one image was generated for each multi-modal prompt via each method, amassing a collection of 300 images. These images were subsequently showcased to 50 participants, who were tasked with indicating their preferences, guided by criteria focused on image and text alignment.
The outcomes, detailed in \cref{table_human_preference_study}, unveil a marked preference for our methods concerning both image and text alignment. It is worth noting that there is a notable majority favoring our methods for image alignment, with over 70\% of the participants expressing a preference for our methods. 

\subsection{Ablation Study}
\label{sec:ablation-study}

To highlight the critical roles of main object description extraction and concept customization via finetuning in our framework, we conducted an ablation study utilizing two alternative strategies: extraction-directly and finetuning-directly. The extraction-directly technique employs the extracted descriptions of the main object to replace images in the multi-modal prompt, thereby creating a pure text prompt. This prompt is then directly applied to a pretrained diffusion model without any finetuning. On the other hand, the finetuning-directly approach skips the step of extracting main object descriptions, opting instead to use special tokens to represent the knowledge of given images throughout the finetuning process. This method generates images based on a pure text prompt formed by substituting images with special tokens.

\begin{table}[tp]
  \centering
  \resizebox{1\columnwidth}{!} {
  \begin{tabular}{c|c|c}
    \toprule
     & Ours-Dreambooth & Ours-Custom Diffusion \\
    \midrule
    Image Alignment & 76.53\% & 71.96\% \\
    \midrule
    Text Alignment & 57.60\% & 51.12\% \\
    \bottomrule
  \end{tabular}
  }
  \caption{\textbf{Human preference study.} This table displays the preferences of participants towards our methods in terms of image alignment and text alignment. The percentages in the table reflect the proportion of participants preferring our methods over the original ones. For instance, Ours-Dreambooth achieved a 57.60\% preference rate for text alignment against Dreambooth, and Ours-Custom Diffusion secured a 71.96\% preference for image alignment compared to Custom Diffusion.
  }
  
  \label{table_human_preference_study}
\end{table}

The results, illustrated in \cref{fig_expr} and quantified by DINO score, CLIP-I score, and CLIP-T score in \cref{table_expr}, demonstrate the limitations of both ablation methods in customization tasks. As detailed in \cref{fig_expr}, neither the extraction-directly nor the finetuning-directly method meets the desired standards for customization. Specifically, the extraction-directly method tends to produce images that resemble the intended concept but fail to achieve precise customization, due to the pure text descriptions' inability to fully capture the given images' essence. The finetuning-directly method exhibits inconsistency in recognizing the object to be customized.

As noted in \cref{table_expr}, aside from the CLIP-T score of the extraction-directly method, all metrics for both ablation methods significantly lag behind those of Ours-Dreambooth and Ours-Custom Diffusion. The high CLIP-T score for the extraction-directly method, despite its customization shortcomings, suggests that while the direct use of extracted descriptions aids in text alignment, it falls at achieving the purpose of accurate object customization.

\section{Conclusion}



We introduce a novel paradigm for customized text-to-image generation that leverages multi-modal prompts as input, refining and expanding upon existing fine-tune-based approaches. This paradigm is distinctly more user-friendly, significantly reducing the amount of information required from users, while also enhancing the capability for detailed customization of complex objects.

Our proposed framework accommodates scenarios involving multiple images within a multi-modal prompt. However, due to limitations in the current stable diffusion models, the performance in multi-image contexts has not reached optimal levels. The underlying reasons for this are further elaborated in the supplementary material. Future efforts will explore the utilization of more advanced diffusion models, such as SDXL \cite{SDXL}, to evaluate our paradigm's effectiveness in multi-image scenarios and to identify strategies for performance enhancement.
Additionally, our current definition of a multi-modal prompt imposes certain constraints, focusing specifically on customizing the main objects within images. Future research will aim to broaden the scope of multi-modal prompts, endeavoring to holistically comprehend and analyze the semantic content encapsulated in both the visual and textual components of the prompt. This expanded interpretation will facilitate a more generalized application of multi-modal prompts, potentially unlocking new avenues for comprehensive image customization.


\bibliographystyle{ACM-Reference-Format}
\bibliography{egbib}


\begin{thebibliography}{82}


\ifx \showCODEN    \undefined \def \showCODEN     #1{\unskip}     \fi
\ifx \showDOI      \undefined \def \showDOI       #1{#1}\fi
\ifx \showISBNx    \undefined \def \showISBNx     #1{\unskip}     \fi
\ifx \showISBNxiii \undefined \def \showISBNxiii  #1{\unskip}     \fi
\ifx \showISSN     \undefined \def \showISSN      #1{\unskip}     \fi
\ifx \showLCCN     \undefined \def \showLCCN      #1{\unskip}     \fi
\ifx \shownote     \undefined \def \shownote      #1{#1}          \fi
\ifx \showarticletitle \undefined \def \showarticletitle #1{#1}   \fi
\ifx \showURL      \undefined \def \showURL       {\relax}        \fi
\providecommand\bibfield[2]{#2}
\providecommand\bibinfo[2]{#2}
\providecommand\natexlab[1]{#1}
\providecommand\showeprint[2][]{arXiv:#2}

\bibitem[Arar et~al\mbox{.}(2023)]%
        {domain-agnostic}
\bibfield{author}{\bibinfo{person}{Moab Arar}, \bibinfo{person}{Rinon Gal}, \bibinfo{person}{Yuval Atzmon}, \bibinfo{person}{Gal Chechik}, \bibinfo{person}{Daniel Cohen-Or}, \bibinfo{person}{Ariel Shamir}, {and} \bibinfo{person}{Amit H.~Bermano}.} \bibinfo{year}{2023}\natexlab{}.
\newblock \showarticletitle{Domain-agnostic tuning-encoder for fast personalization of text-to-image models}. In \bibinfo{booktitle}{\emph{SIGGRAPH Asia 2023 Conference Papers}}. \bibinfo{pages}{1--10}.
\newblock


\bibitem[Avrahami et~al\mbox{.}(2023a)]%
        {break_a_scene}
\bibfield{author}{\bibinfo{person}{Omri Avrahami}, \bibinfo{person}{Kfir Aberman}, \bibinfo{person}{Ohad Fried}, \bibinfo{person}{Daniel Cohen-Or}, {and} \bibinfo{person}{Dani Lischinski}.} \bibinfo{year}{2023}\natexlab{a}.
\newblock \showarticletitle{Break-A-Scene: Extracting Multiple Concepts from a Single Image}.
\newblock \bibinfo{journal}{\emph{arXiv preprint arXiv:2305.16311}} (\bibinfo{year}{2023}).
\newblock


\bibitem[Avrahami et~al\mbox{.}(2023b)]%
        {spatext}
\bibfield{author}{\bibinfo{person}{Omri Avrahami}, \bibinfo{person}{Thomas Hayes}, \bibinfo{person}{Oran Gafni}, \bibinfo{person}{Sonal Gupta}, \bibinfo{person}{Yaniv Taigman}, \bibinfo{person}{Devi Parikh}, \bibinfo{person}{Dani Lischinski}, \bibinfo{person}{Ohad Fried}, {and} \bibinfo{person}{Xi Yin}.} \bibinfo{year}{2023}\natexlab{b}.
\newblock \showarticletitle{Spatext: Spatio-textual representation for controllable image generation}. In \bibinfo{booktitle}{\emph{Proceedings of the IEEE/CVF Conference on Computer Vision and Pattern Recognition}}. \bibinfo{pages}{18370--18380}.
\newblock


\bibitem[Balaji et~al\mbox{.}(2022)]%
        {ediffi}
\bibfield{author}{\bibinfo{person}{Yogesh Balaji}, \bibinfo{person}{Seungjun Nah}, \bibinfo{person}{Xun Huang}, \bibinfo{person}{Arash Vahdat}, \bibinfo{person}{Jiaming Song}, \bibinfo{person}{Karsten Kreis}, \bibinfo{person}{Miika Aittala}, \bibinfo{person}{Timo Aila}, \bibinfo{person}{Samuli Laine}, \bibinfo{person}{Bryan Catanzaro}, {et~al\mbox{.}}} \bibinfo{year}{2022}\natexlab{}.
\newblock \showarticletitle{ediffi: Text-to-image diffusion models with an ensemble of expert denoisers}.
\newblock \bibinfo{journal}{\emph{arXiv preprint arXiv:2211.01324}} (\bibinfo{year}{2022}).
\newblock


\bibitem[Blattmann et~al\mbox{.}(2023)]%
        {align_your_latents}
\bibfield{author}{\bibinfo{person}{Andreas Blattmann}, \bibinfo{person}{Robin Rombach}, \bibinfo{person}{Huan Ling}, \bibinfo{person}{Tim Dockhorn}, \bibinfo{person}{Seung~Wook Kim}, \bibinfo{person}{Sanja Fidler}, {and} \bibinfo{person}{Karsten Kreis}.} \bibinfo{year}{2023}\natexlab{}.
\newblock \showarticletitle{Align your latents: High-resolution video synthesis with latent diffusion models}. In \bibinfo{booktitle}{\emph{Proceedings of the IEEE/CVF Conference on Computer Vision and Pattern Recognition}}. \bibinfo{pages}{22563--22575}.
\newblock


\bibitem[Brooks et~al\mbox{.}(2023)]%
        {instructpix2pix}
\bibfield{author}{\bibinfo{person}{Tim Brooks}, \bibinfo{person}{Aleksander Holynski}, {and} \bibinfo{person}{Alexei~A Efros}.} \bibinfo{year}{2023}\natexlab{}.
\newblock \showarticletitle{Instructpix2pix: Learning to follow image editing instructions}. In \bibinfo{booktitle}{\emph{Proceedings of the IEEE/CVF Conference on Computer Vision and Pattern Recognition}}. \bibinfo{pages}{18392--18402}.
\newblock


\bibitem[Brown et~al\mbox{.}(2020)]%
        {GPT3}
\bibfield{author}{\bibinfo{person}{Tom Brown}, \bibinfo{person}{Benjamin Mann}, \bibinfo{person}{Nick Ryder}, \bibinfo{person}{Melanie Subbiah}, \bibinfo{person}{Jared~D Kaplan}, \bibinfo{person}{Prafulla Dhariwal}, \bibinfo{person}{Arvind Neelakantan}, \bibinfo{person}{Pranav Shyam}, \bibinfo{person}{Girish Sastry}, \bibinfo{person}{Amanda Askell}, {et~al\mbox{.}}} \bibinfo{year}{2020}\natexlab{}.
\newblock \showarticletitle{Language models are few-shot learners}.
\newblock \bibinfo{journal}{\emph{Advances in neural information processing systems}}  \bibinfo{volume}{33} (\bibinfo{year}{2020}), \bibinfo{pages}{1877--1901}.
\newblock


\bibitem[Caron et~al\mbox{.}(2021)]%
        {dino}
\bibfield{author}{\bibinfo{person}{Mathilde Caron}, \bibinfo{person}{Hugo Touvron}, \bibinfo{person}{Ishan Misra}, \bibinfo{person}{Herv{\'e} J{\'e}gou}, \bibinfo{person}{Julien Mairal}, \bibinfo{person}{Piotr Bojanowski}, {and} \bibinfo{person}{Armand Joulin}.} \bibinfo{year}{2021}\natexlab{}.
\newblock \showarticletitle{Emerging properties in self-supervised vision transformers}. In \bibinfo{booktitle}{\emph{Proceedings of the IEEE/CVF international conference on computer vision}}. \bibinfo{pages}{9650--9660}.
\newblock


\bibitem[Chai et~al\mbox{.}(2023)]%
        {stablevideo}
\bibfield{author}{\bibinfo{person}{Wenhao Chai}, \bibinfo{person}{Xun Guo}, \bibinfo{person}{Gaoang Wang}, {and} \bibinfo{person}{Yan Lu}.} \bibinfo{year}{2023}\natexlab{}.
\newblock \showarticletitle{Stablevideo: Text-driven consistency-aware diffusion video editing}. In \bibinfo{booktitle}{\emph{Proceedings of the IEEE/CVF International Conference on Computer Vision}}. \bibinfo{pages}{23040--23050}.
\newblock


\bibitem[Chang et~al\mbox{.}(2023)]%
        {muse}
\bibfield{author}{\bibinfo{person}{Huiwen Chang}, \bibinfo{person}{Han Zhang}, \bibinfo{person}{Jarred Barber}, \bibinfo{person}{AJ Maschinot}, \bibinfo{person}{Jose Lezama}, \bibinfo{person}{Lu Jiang}, \bibinfo{person}{Ming-Hsuan Yang}, \bibinfo{person}{Kevin Murphy}, \bibinfo{person}{William~T Freeman}, \bibinfo{person}{Michael Rubinstein}, {et~al\mbox{.}}} \bibinfo{year}{2023}\natexlab{}.
\newblock \showarticletitle{Muse: Text-to-image generation via masked generative transformers}.
\newblock \bibinfo{journal}{\emph{arXiv preprint arXiv:2301.00704}} (\bibinfo{year}{2023}).
\newblock


\bibitem[Chefer et~al\mbox{.}(2023)]%
        {attend_and_excite}
\bibfield{author}{\bibinfo{person}{Hila Chefer}, \bibinfo{person}{Yuval Alaluf}, \bibinfo{person}{Yael Vinker}, \bibinfo{person}{Lior Wolf}, {and} \bibinfo{person}{Daniel Cohen-Or}.} \bibinfo{year}{2023}\natexlab{}.
\newblock \showarticletitle{Attend-and-excite: Attention-based semantic guidance for text-to-image diffusion models}.
\newblock \bibinfo{journal}{\emph{ACM Transactions on Graphics (TOG)}} \bibinfo{volume}{42}, \bibinfo{number}{4} (\bibinfo{year}{2023}), \bibinfo{pages}{1--10}.
\newblock


\bibitem[Chen and Shu(2023a)]%
        {misinformation2}
\bibfield{author}{\bibinfo{person}{Canyu Chen} {and} \bibinfo{person}{Kai Shu}.} \bibinfo{year}{2023}\natexlab{a}.
\newblock \showarticletitle{Can llm-generated misinformation be detected?}
\newblock \bibinfo{journal}{\emph{arXiv preprint arXiv:2309.13788}} (\bibinfo{year}{2023}).
\newblock


\bibitem[Chen and Shu(2023b)]%
        {misinformation}
\bibfield{author}{\bibinfo{person}{Canyu Chen} {and} \bibinfo{person}{Kai Shu}.} \bibinfo{year}{2023}\natexlab{b}.
\newblock \showarticletitle{Combating misinformation in the age of llms: Opportunities and challenges}.
\newblock \bibinfo{journal}{\emph{arXiv preprint arXiv:2311.05656}} (\bibinfo{year}{2023}).
\newblock


\bibitem[Chen et~al\mbox{.}(2023a)]%
        {X_llm}
\bibfield{author}{\bibinfo{person}{Feilong Chen}, \bibinfo{person}{Minglun Han}, \bibinfo{person}{Haozhi Zhao}, \bibinfo{person}{Qingyang Zhang}, \bibinfo{person}{Jing Shi}, \bibinfo{person}{Shuang Xu}, {and} \bibinfo{person}{Bo Xu}.} \bibinfo{year}{2023}\natexlab{a}.
\newblock \showarticletitle{X-llm: Bootstrapping advanced large language models by treating multi-modalities as foreign languages}.
\newblock \bibinfo{journal}{\emph{arXiv preprint arXiv:2305.04160}} (\bibinfo{year}{2023}).
\newblock


\bibitem[Chen et~al\mbox{.}(2023b)]%
        {disenbooth}
\bibfield{author}{\bibinfo{person}{Hong Chen}, \bibinfo{person}{Yipeng Zhang}, \bibinfo{person}{Xin Wang}, \bibinfo{person}{Xuguang Duan}, \bibinfo{person}{Yuwei Zhou}, {and} \bibinfo{person}{Wenwu Zhu}.} \bibinfo{year}{2023}\natexlab{b}.
\newblock \showarticletitle{DisenBooth: Disentangled Parameter-Efficient Tuning for Subject-Driven Text-to-Image Generation}.
\newblock \bibinfo{journal}{\emph{arXiv preprint arXiv:2305.03374}} (\bibinfo{year}{2023}).
\newblock


\bibitem[Chen et~al\mbox{.}(2023c)]%
        {shikra}
\bibfield{author}{\bibinfo{person}{Keqin Chen}, \bibinfo{person}{Zhao Zhang}, \bibinfo{person}{Weili Zeng}, \bibinfo{person}{Richong Zhang}, \bibinfo{person}{Feng Zhu}, {and} \bibinfo{person}{Rui Zhao}.} \bibinfo{year}{2023}\natexlab{c}.
\newblock \showarticletitle{Shikra: Unleashing Multimodal LLM's Referential Dialogue Magic}.
\newblock \bibinfo{journal}{\emph{arXiv preprint arXiv:2306.15195}} (\bibinfo{year}{2023}).
\newblock


\bibitem[Chen et~al\mbox{.}(2024b)]%
        {cross_attention_guidance}
\bibfield{author}{\bibinfo{person}{Minghao Chen}, \bibinfo{person}{Iro Laina}, {and} \bibinfo{person}{Andrea Vedaldi}.} \bibinfo{year}{2024}\natexlab{b}.
\newblock \showarticletitle{Training-free layout control with cross-attention guidance}. In \bibinfo{booktitle}{\emph{Proceedings of the IEEE/CVF Winter Conference on Applications of Computer Vision}}. \bibinfo{pages}{5343--5353}.
\newblock


\bibitem[Chen et~al\mbox{.}(2024a)]%
        {subject_driven_generation}
\bibfield{author}{\bibinfo{person}{Wenhu Chen}, \bibinfo{person}{Hexiang Hu}, \bibinfo{person}{Yandong Li}, \bibinfo{person}{Nataniel Ruiz}, \bibinfo{person}{Xuhui Jia}, \bibinfo{person}{Ming-Wei Chang}, {and} \bibinfo{person}{William~W Cohen}.} \bibinfo{year}{2024}\natexlab{a}.
\newblock \showarticletitle{Subject-driven text-to-image generation via apprenticeship learning}.
\newblock \bibinfo{journal}{\emph{Advances in Neural Information Processing Systems}}  \bibinfo{volume}{36} (\bibinfo{year}{2024}).
\newblock


\bibitem[Devlin et~al\mbox{.}(2018)]%
        {BERT}
\bibfield{author}{\bibinfo{person}{Jacob Devlin}, \bibinfo{person}{Ming-Wei Chang}, \bibinfo{person}{Kenton Lee}, {and} \bibinfo{person}{Kristina Toutanova}.} \bibinfo{year}{2018}\natexlab{}.
\newblock \showarticletitle{Bert: Pre-training of deep bidirectional transformers for language understanding}.
\newblock \bibinfo{journal}{\emph{arXiv preprint arXiv:1810.04805}} (\bibinfo{year}{2018}).
\newblock


\bibitem[Epstein et~al\mbox{.}(2024)]%
        {controllable_generation}
\bibfield{author}{\bibinfo{person}{Dave Epstein}, \bibinfo{person}{Allan Jabri}, \bibinfo{person}{Ben Poole}, \bibinfo{person}{Alexei Efros}, {and} \bibinfo{person}{Aleksander Holynski}.} \bibinfo{year}{2024}\natexlab{}.
\newblock \showarticletitle{Diffusion self-guidance for controllable image generation}.
\newblock \bibinfo{journal}{\emph{Advances in Neural Information Processing Systems}}  \bibinfo{volume}{36} (\bibinfo{year}{2024}).
\newblock


\bibitem[Esser et~al\mbox{.}(2023)]%
        {video_synthesis}
\bibfield{author}{\bibinfo{person}{Patrick Esser}, \bibinfo{person}{Johnathan Chiu}, \bibinfo{person}{Parmida Atighehchian}, \bibinfo{person}{Jonathan Granskog}, {and} \bibinfo{person}{Anastasis Germanidis}.} \bibinfo{year}{2023}\natexlab{}.
\newblock \showarticletitle{Structure and content-guided video synthesis with diffusion models}. In \bibinfo{booktitle}{\emph{Proceedings of the IEEE/CVF International Conference on Computer Vision}}. \bibinfo{pages}{7346--7356}.
\newblock


\bibitem[Farhadi et~al\mbox{.}(2010)]%
        {image-captioning-template1}
\bibfield{author}{\bibinfo{person}{Ali Farhadi}, \bibinfo{person}{Mohsen Hejrati}, \bibinfo{person}{Mohammad~Amin Sadeghi}, \bibinfo{person}{Peter Young}, \bibinfo{person}{Cyrus Rashtchian}, \bibinfo{person}{Julia Hockenmaier}, {and} \bibinfo{person}{David Forsyth}.} \bibinfo{year}{2010}\natexlab{}.
\newblock \showarticletitle{Every picture tells a story: Generating sentences from images}. In \bibinfo{booktitle}{\emph{Computer Vision--ECCV 2010: 11th European Conference on Computer Vision, Heraklion, Crete, Greece, September 5-11, 2010, Proceedings, Part IV 11}}. Springer, \bibinfo{pages}{15--29}.
\newblock


\bibitem[Gal et~al\mbox{.}(2022)]%
        {textual-inversion}
\bibfield{author}{\bibinfo{person}{Rinon Gal}, \bibinfo{person}{Yuval Alaluf}, \bibinfo{person}{Yuval Atzmon}, \bibinfo{person}{Or Patashnik}, \bibinfo{person}{Amit~H Bermano}, \bibinfo{person}{Gal Chechik}, {and} \bibinfo{person}{Daniel Cohen-Or}.} \bibinfo{year}{2022}\natexlab{}.
\newblock \showarticletitle{An image is worth one word: Personalizing text-to-image generation using textual inversion}.
\newblock \bibinfo{journal}{\emph{arXiv preprint arXiv:2208.01618}} (\bibinfo{year}{2022}).
\newblock


\bibitem[Gal et~al\mbox{.}(2023)]%
        {fast_personalization}
\bibfield{author}{\bibinfo{person}{Rinon Gal}, \bibinfo{person}{Moab Arar}, \bibinfo{person}{Yuval Atzmon}, \bibinfo{person}{Amit~H Bermano}, \bibinfo{person}{Gal Chechik}, {and} \bibinfo{person}{Daniel Cohen-Or}.} \bibinfo{year}{2023}\natexlab{}.
\newblock \showarticletitle{Encoder-based domain tuning for fast personalization of text-to-image models}.
\newblock \bibinfo{journal}{\emph{ACM Transactions on Graphics (TOG)}} \bibinfo{volume}{42}, \bibinfo{number}{4} (\bibinfo{year}{2023}), \bibinfo{pages}{1--13}.
\newblock


\bibitem[Ge et~al\mbox{.}(2023)]%
        {video_preserve}
\bibfield{author}{\bibinfo{person}{Songwei Ge}, \bibinfo{person}{Seungjun Nah}, \bibinfo{person}{Guilin Liu}, \bibinfo{person}{Tyler Poon}, \bibinfo{person}{Andrew Tao}, \bibinfo{person}{Bryan Catanzaro}, \bibinfo{person}{David Jacobs}, \bibinfo{person}{Jia-Bin Huang}, \bibinfo{person}{Ming-Yu Liu}, {and} \bibinfo{person}{Yogesh Balaji}.} \bibinfo{year}{2023}\natexlab{}.
\newblock \showarticletitle{Preserve your own correlation: A noise prior for video diffusion models}. In \bibinfo{booktitle}{\emph{Proceedings of the IEEE/CVF International Conference on Computer Vision}}. \bibinfo{pages}{22930--22941}.
\newblock


\bibitem[Gong et~al\mbox{.}(2014)]%
        {image-captioning-retrieval1}
\bibfield{author}{\bibinfo{person}{Yunchao Gong}, \bibinfo{person}{Liwei Wang}, \bibinfo{person}{Micah Hodosh}, \bibinfo{person}{Julia Hockenmaier}, {and} \bibinfo{person}{Svetlana Lazebnik}.} \bibinfo{year}{2014}\natexlab{}.
\newblock \showarticletitle{Improving image-sentence embeddings using large weakly annotated photo collections}. In \bibinfo{booktitle}{\emph{Computer Vision--ECCV 2014: 13th European Conference, Zurich, Switzerland, September 6-12, 2014, Proceedings, Part IV 13}}. Springer, \bibinfo{pages}{529--545}.
\newblock


\bibitem[Goodfellow et~al\mbox{.}(2020)]%
        {GAN}
\bibfield{author}{\bibinfo{person}{Ian Goodfellow}, \bibinfo{person}{Jean Pouget-Abadie}, \bibinfo{person}{Mehdi Mirza}, \bibinfo{person}{Bing Xu}, \bibinfo{person}{David Warde-Farley}, \bibinfo{person}{Sherjil Ozair}, \bibinfo{person}{Aaron Courville}, {and} \bibinfo{person}{Yoshua Bengio}.} \bibinfo{year}{2020}\natexlab{}.
\newblock \showarticletitle{Generative adversarial networks}.
\newblock \bibinfo{journal}{\emph{Commun. ACM}} \bibinfo{volume}{63}, \bibinfo{number}{11} (\bibinfo{year}{2020}), \bibinfo{pages}{139--144}.
\newblock


\bibitem[Gu et~al\mbox{.}(2022)]%
        {vector_quantized_diffusion}
\bibfield{author}{\bibinfo{person}{Shuyang Gu}, \bibinfo{person}{Dong Chen}, \bibinfo{person}{Jianmin Bao}, \bibinfo{person}{Fang Wen}, \bibinfo{person}{Bo Zhang}, \bibinfo{person}{Dongdong Chen}, \bibinfo{person}{Lu Yuan}, {and} \bibinfo{person}{Baining Guo}.} \bibinfo{year}{2022}\natexlab{}.
\newblock \showarticletitle{Vector quantized diffusion model for text-to-image synthesis}. In \bibinfo{booktitle}{\emph{Proceedings of the IEEE/CVF Conference on Computer Vision and Pattern Recognition}}. \bibinfo{pages}{10696--10706}.
\newblock


\bibitem[Gu et~al\mbox{.}(2024)]%
        {mix_of_show}
\bibfield{author}{\bibinfo{person}{Yuchao Gu}, \bibinfo{person}{Xintao Wang}, \bibinfo{person}{Jay~Zhangjie Wu}, \bibinfo{person}{Yujun Shi}, \bibinfo{person}{Yunpeng Chen}, \bibinfo{person}{Zihan Fan}, \bibinfo{person}{Wuyou Xiao}, \bibinfo{person}{Rui Zhao}, \bibinfo{person}{Shuning Chang}, \bibinfo{person}{Weijia Wu}, {et~al\mbox{.}}} \bibinfo{year}{2024}\natexlab{}.
\newblock \showarticletitle{Mix-of-show: Decentralized low-rank adaptation for multi-concept customization of diffusion models}.
\newblock \bibinfo{journal}{\emph{Advances in Neural Information Processing Systems}}  \bibinfo{volume}{36} (\bibinfo{year}{2024}).
\newblock


\bibitem[Guo et~al\mbox{.}(2023)]%
        {animatediff}
\bibfield{author}{\bibinfo{person}{Yuwei Guo}, \bibinfo{person}{Ceyuan Yang}, \bibinfo{person}{Anyi Rao}, \bibinfo{person}{Yaohui Wang}, \bibinfo{person}{Yu Qiao}, \bibinfo{person}{Dahua Lin}, {and} \bibinfo{person}{Bo Dai}.} \bibinfo{year}{2023}\natexlab{}.
\newblock \showarticletitle{Animatediff: Animate your personalized text-to-image diffusion models without specific tuning}.
\newblock \bibinfo{journal}{\emph{arXiv preprint arXiv:2307.04725}} (\bibinfo{year}{2023}).
\newblock


\bibitem[Hao et~al\mbox{.}(2023)]%
        {vico}
\bibfield{author}{\bibinfo{person}{Shaozhe Hao}, \bibinfo{person}{Kai Han}, \bibinfo{person}{Shihao Zhao}, {and} \bibinfo{person}{Kwan-Yee~K Wong}.} \bibinfo{year}{2023}\natexlab{}.
\newblock \showarticletitle{Vico: Detail-preserving visual condition for personalized text-to-image generation}.
\newblock \bibinfo{journal}{\emph{arXiv preprint arXiv:2306.00971}} (\bibinfo{year}{2023}).
\newblock


\bibitem[Ho et~al\mbox{.}(2020)]%
        {DDPM}
\bibfield{author}{\bibinfo{person}{Jonathan Ho}, \bibinfo{person}{Ajay Jain}, {and} \bibinfo{person}{Pieter Abbeel}.} \bibinfo{year}{2020}\natexlab{}.
\newblock \showarticletitle{Denoising diffusion probabilistic models}.
\newblock \bibinfo{journal}{\emph{Advances in neural information processing systems}}  \bibinfo{volume}{33} (\bibinfo{year}{2020}), \bibinfo{pages}{6840--6851}.
\newblock


\bibitem[Hodosh et~al\mbox{.}(2013)]%
        {image-captioning-retrieval2}
\bibfield{author}{\bibinfo{person}{Micah Hodosh}, \bibinfo{person}{Peter Young}, {and} \bibinfo{person}{Julia Hockenmaier}.} \bibinfo{year}{2013}\natexlab{}.
\newblock \showarticletitle{Framing image description as a ranking task: Data, models and evaluation metrics}.
\newblock \bibinfo{journal}{\emph{Journal of Artificial Intelligence Research}}  \bibinfo{volume}{47} (\bibinfo{year}{2013}), \bibinfo{pages}{853--899}.
\newblock


\bibitem[Kawar et~al\mbox{.}(2023)]%
        {imagic}
\bibfield{author}{\bibinfo{person}{Bahjat Kawar}, \bibinfo{person}{Shiran Zada}, \bibinfo{person}{Oran Lang}, \bibinfo{person}{Omer Tov}, \bibinfo{person}{Huiwen Chang}, \bibinfo{person}{Tali Dekel}, \bibinfo{person}{Inbar Mosseri}, {and} \bibinfo{person}{Michal Irani}.} \bibinfo{year}{2023}\natexlab{}.
\newblock \showarticletitle{Imagic: Text-based real image editing with diffusion models}. In \bibinfo{booktitle}{\emph{Proceedings of the IEEE/CVF Conference on Computer Vision and Pattern Recognition}}. \bibinfo{pages}{6007--6017}.
\newblock


\bibitem[Koh et~al\mbox{.}(2024)]%
        {multimodal_llm2}
\bibfield{author}{\bibinfo{person}{Jing~Yu Koh}, \bibinfo{person}{Daniel Fried}, {and} \bibinfo{person}{Russ~R Salakhutdinov}.} \bibinfo{year}{2024}\natexlab{}.
\newblock \showarticletitle{Generating images with multimodal language models}.
\newblock \bibinfo{journal}{\emph{Advances in Neural Information Processing Systems}}  \bibinfo{volume}{36} (\bibinfo{year}{2024}).
\newblock


\bibitem[Kumari et~al\mbox{.}(2023a)]%
        {ablation_concept}
\bibfield{author}{\bibinfo{person}{Nupur Kumari}, \bibinfo{person}{Bingliang Zhang}, \bibinfo{person}{Sheng-Yu Wang}, \bibinfo{person}{Eli Shechtman}, \bibinfo{person}{Richard Zhang}, {and} \bibinfo{person}{Jun-Yan Zhu}.} \bibinfo{year}{2023}\natexlab{a}.
\newblock \showarticletitle{Ablating concepts in text-to-image diffusion models}. In \bibinfo{booktitle}{\emph{Proceedings of the IEEE/CVF International Conference on Computer Vision}}. \bibinfo{pages}{22691--22702}.
\newblock


\bibitem[Kumari et~al\mbox{.}(2023b)]%
        {custom-diffusion}
\bibfield{author}{\bibinfo{person}{Nupur Kumari}, \bibinfo{person}{Bingliang Zhang}, \bibinfo{person}{Richard Zhang}, \bibinfo{person}{Eli Shechtman}, {and} \bibinfo{person}{Jun-Yan Zhu}.} \bibinfo{year}{2023}\natexlab{b}.
\newblock \showarticletitle{Multi-concept customization of text-to-image diffusion}. In \bibinfo{booktitle}{\emph{Proceedings of the IEEE/CVF Conference on Computer Vision and Pattern Recognition}}. \bibinfo{pages}{1931--1941}.
\newblock


\bibitem[Li et~al\mbox{.}(2024)]%
        {blip_diffusion}
\bibfield{author}{\bibinfo{person}{Dongxu Li}, \bibinfo{person}{Junnan Li}, {and} \bibinfo{person}{Steven Hoi}.} \bibinfo{year}{2024}\natexlab{}.
\newblock \showarticletitle{Blip-diffusion: Pre-trained subject representation for controllable text-to-image generation and editing}.
\newblock \bibinfo{journal}{\emph{Advances in Neural Information Processing Systems}}  \bibinfo{volume}{36} (\bibinfo{year}{2024}).
\newblock


\bibitem[Li et~al\mbox{.}(2022)]%
        {BLIP}
\bibfield{author}{\bibinfo{person}{Junnan Li}, \bibinfo{person}{Dongxu Li}, \bibinfo{person}{Caiming Xiong}, {and} \bibinfo{person}{Steven Hoi}.} \bibinfo{year}{2022}\natexlab{}.
\newblock \showarticletitle{Blip: Bootstrapping language-image pre-training for unified vision-language understanding and generation}. In \bibinfo{booktitle}{\emph{International Conference on Machine Learning}}. PMLR, \bibinfo{pages}{12888--12900}.
\newblock


\bibitem[Li et~al\mbox{.}(2011)]%
        {image-captioning-template2}
\bibfield{author}{\bibinfo{person}{Siming Li}, \bibinfo{person}{Girish Kulkarni}, \bibinfo{person}{Tamara Berg}, \bibinfo{person}{Alexander Berg}, {and} \bibinfo{person}{Yejin Choi}.} \bibinfo{year}{2011}\natexlab{}.
\newblock \showarticletitle{Composing simple image descriptions using web-scale n-grams}. In \bibinfo{booktitle}{\emph{Proceedings of the fifteenth conference on computational natural language learning}}. \bibinfo{pages}{220--228}.
\newblock


\bibitem[Lin et~al\mbox{.}(2023)]%
        {magic3d}
\bibfield{author}{\bibinfo{person}{Chen-Hsuan Lin}, \bibinfo{person}{Jun Gao}, \bibinfo{person}{Luming Tang}, \bibinfo{person}{Towaki Takikawa}, \bibinfo{person}{Xiaohui Zeng}, \bibinfo{person}{Xun Huang}, \bibinfo{person}{Karsten Kreis}, \bibinfo{person}{Sanja Fidler}, \bibinfo{person}{Ming-Yu Liu}, {and} \bibinfo{person}{Tsung-Yi Lin}.} \bibinfo{year}{2023}\natexlab{}.
\newblock \showarticletitle{Magic3d: High-resolution text-to-3d content creation}. In \bibinfo{booktitle}{\emph{Proceedings of the IEEE/CVF Conference on Computer Vision and Pattern Recognition}}. \bibinfo{pages}{300--309}.
\newblock


\bibitem[Liu et~al\mbox{.}(2023)]%
        {cones}
\bibfield{author}{\bibinfo{person}{Zhiheng Liu}, \bibinfo{person}{Ruili Feng}, \bibinfo{person}{Kai Zhu}, \bibinfo{person}{Yifei Zhang}, \bibinfo{person}{Kecheng Zheng}, \bibinfo{person}{Yu Liu}, \bibinfo{person}{Deli Zhao}, \bibinfo{person}{Jingren Zhou}, {and} \bibinfo{person}{Yang Cao}.} \bibinfo{year}{2023}\natexlab{}.
\newblock \showarticletitle{Cones: Concept neurons in diffusion models for customized generation}.
\newblock \bibinfo{journal}{\emph{arXiv preprint arXiv:2303.05125}} (\bibinfo{year}{2023}).
\newblock


\bibitem[Mokady et~al\mbox{.}(2023)]%
        {editing_inversion}
\bibfield{author}{\bibinfo{person}{Ron Mokady}, \bibinfo{person}{Amir Hertz}, \bibinfo{person}{Kfir Aberman}, \bibinfo{person}{Yael Pritch}, {and} \bibinfo{person}{Daniel Cohen-Or}.} \bibinfo{year}{2023}\natexlab{}.
\newblock \showarticletitle{Null-text inversion for editing real images using guided diffusion models}. In \bibinfo{booktitle}{\emph{Proceedings of the IEEE/CVF Conference on Computer Vision and Pattern Recognition}}. \bibinfo{pages}{6038--6047}.
\newblock


\bibitem[Nichol and Dhariwal(2021)]%
        {IDDPM}
\bibfield{author}{\bibinfo{person}{Alexander~Quinn Nichol} {and} \bibinfo{person}{Prafulla Dhariwal}.} \bibinfo{year}{2021}\natexlab{}.
\newblock \showarticletitle{Improved denoising diffusion probabilistic models}. In \bibinfo{booktitle}{\emph{International Conference on Machine Learning}}. PMLR, \bibinfo{pages}{8162--8171}.
\newblock


\bibitem[Ordonez et~al\mbox{.}(2011)]%
        {image-captioning-retrieval3}
\bibfield{author}{\bibinfo{person}{Vicente Ordonez}, \bibinfo{person}{Girish Kulkarni}, {and} \bibinfo{person}{Tamara Berg}.} \bibinfo{year}{2011}\natexlab{}.
\newblock \showarticletitle{Im2text: Describing images using 1 million captioned photographs}.
\newblock \bibinfo{journal}{\emph{Advances in neural information processing systems}}  \bibinfo{volume}{24} (\bibinfo{year}{2011}).
\newblock


\bibitem[Ouyang et~al\mbox{.}(2022)]%
        {InstructGPT}
\bibfield{author}{\bibinfo{person}{Long Ouyang}, \bibinfo{person}{Jeffrey Wu}, \bibinfo{person}{Xu Jiang}, \bibinfo{person}{Diogo Almeida}, \bibinfo{person}{Carroll Wainwright}, \bibinfo{person}{Pamela Mishkin}, \bibinfo{person}{Chong Zhang}, \bibinfo{person}{Sandhini Agarwal}, \bibinfo{person}{Katarina Slama}, \bibinfo{person}{Alex Ray}, {et~al\mbox{.}}} \bibinfo{year}{2022}\natexlab{}.
\newblock \showarticletitle{Training language models to follow instructions with human feedback}.
\newblock \bibinfo{journal}{\emph{Advances in Neural Information Processing Systems}}  \bibinfo{volume}{35} (\bibinfo{year}{2022}), \bibinfo{pages}{27730--27744}.
\newblock


\bibitem[Podell et~al\mbox{.}(2023)]%
        {SDXL}
\bibfield{author}{\bibinfo{person}{Dustin Podell}, \bibinfo{person}{Zion English}, \bibinfo{person}{Kyle Lacey}, \bibinfo{person}{Andreas Blattmann}, \bibinfo{person}{Tim Dockhorn}, \bibinfo{person}{Jonas M{\"u}ller}, \bibinfo{person}{Joe Penna}, {and} \bibinfo{person}{Robin Rombach}.} \bibinfo{year}{2023}\natexlab{}.
\newblock \showarticletitle{Sdxl: Improving latent diffusion models for high-resolution image synthesis}.
\newblock \bibinfo{journal}{\emph{arXiv preprint arXiv:2307.01952}} (\bibinfo{year}{2023}).
\newblock


\bibitem[Radford et~al\mbox{.}(2021)]%
        {clip}
\bibfield{author}{\bibinfo{person}{Alec Radford}, \bibinfo{person}{Jong~Wook Kim}, \bibinfo{person}{Chris Hallacy}, \bibinfo{person}{Aditya Ramesh}, \bibinfo{person}{Gabriel Goh}, \bibinfo{person}{Sandhini Agarwal}, \bibinfo{person}{Girish Sastry}, \bibinfo{person}{Amanda Askell}, \bibinfo{person}{Pamela Mishkin}, \bibinfo{person}{Jack Clark}, {et~al\mbox{.}}} \bibinfo{year}{2021}\natexlab{}.
\newblock \showarticletitle{Learning transferable visual models from natural language supervision}. In \bibinfo{booktitle}{\emph{International conference on machine learning}}. PMLR, \bibinfo{pages}{8748--8763}.
\newblock


\bibitem[Raffel et~al\mbox{.}(2020)]%
        {T5}
\bibfield{author}{\bibinfo{person}{Colin Raffel}, \bibinfo{person}{Noam Shazeer}, \bibinfo{person}{Adam Roberts}, \bibinfo{person}{Katherine Lee}, \bibinfo{person}{Sharan Narang}, \bibinfo{person}{Michael Matena}, \bibinfo{person}{Yanqi Zhou}, \bibinfo{person}{Wei Li}, {and} \bibinfo{person}{Peter~J Liu}.} \bibinfo{year}{2020}\natexlab{}.
\newblock \showarticletitle{Exploring the limits of transfer learning with a unified text-to-text transformer}.
\newblock \bibinfo{journal}{\emph{The Journal of Machine Learning Research}} \bibinfo{volume}{21}, \bibinfo{number}{1} (\bibinfo{year}{2020}), \bibinfo{pages}{5485--5551}.
\newblock


\bibitem[Ramesh et~al\mbox{.}(2022)]%
        {DALLE2}
\bibfield{author}{\bibinfo{person}{Aditya Ramesh}, \bibinfo{person}{Prafulla Dhariwal}, \bibinfo{person}{Alex Nichol}, \bibinfo{person}{Casey Chu}, {and} \bibinfo{person}{Mark Chen}.} \bibinfo{year}{2022}\natexlab{}.
\newblock \showarticletitle{Hierarchical text-conditional image generation with clip latents}.
\newblock \bibinfo{journal}{\emph{arXiv preprint arXiv:2204.06125}} \bibinfo{volume}{1}, \bibinfo{number}{2} (\bibinfo{year}{2022}), \bibinfo{pages}{3}.
\newblock


\bibitem[Reed et~al\mbox{.}(2016)]%
        {Text-Conditional-GAN}
\bibfield{author}{\bibinfo{person}{Scott Reed}, \bibinfo{person}{Zeynep Akata}, \bibinfo{person}{Xinchen Yan}, \bibinfo{person}{Lajanugen Logeswaran}, \bibinfo{person}{Bernt Schiele}, {and} \bibinfo{person}{Honglak Lee}.} \bibinfo{year}{2016}\natexlab{}.
\newblock \showarticletitle{Generative adversarial text to image synthesis}. In \bibinfo{booktitle}{\emph{International conference on machine learning}}. PMLR, \bibinfo{pages}{1060--1069}.
\newblock


\bibitem[Rombach et~al\mbox{.}(2022)]%
        {SD}
\bibfield{author}{\bibinfo{person}{Robin Rombach}, \bibinfo{person}{Andreas Blattmann}, \bibinfo{person}{Dominik Lorenz}, \bibinfo{person}{Patrick Esser}, {and} \bibinfo{person}{Bj{\"o}rn Ommer}.} \bibinfo{year}{2022}\natexlab{}.
\newblock \showarticletitle{High-resolution image synthesis with latent diffusion models}. In \bibinfo{booktitle}{\emph{Proceedings of the IEEE/CVF conference on computer vision and pattern recognition}}. \bibinfo{pages}{10684--10695}.
\newblock


\bibitem[Ruiz et~al\mbox{.}(2023a)]%
        {dreambooth}
\bibfield{author}{\bibinfo{person}{Nataniel Ruiz}, \bibinfo{person}{Yuanzhen Li}, \bibinfo{person}{Varun Jampani}, \bibinfo{person}{Yael Pritch}, \bibinfo{person}{Michael Rubinstein}, {and} \bibinfo{person}{Kfir Aberman}.} \bibinfo{year}{2023}\natexlab{a}.
\newblock \showarticletitle{Dreambooth: Fine tuning text-to-image diffusion models for subject-driven generation}. In \bibinfo{booktitle}{\emph{Proceedings of the IEEE/CVF Conference on Computer Vision and Pattern Recognition}}. \bibinfo{pages}{22500--22510}.
\newblock


\bibitem[Ruiz et~al\mbox{.}(2023b)]%
        {hyperdreambooth}
\bibfield{author}{\bibinfo{person}{Nataniel Ruiz}, \bibinfo{person}{Yuanzhen Li}, \bibinfo{person}{Varun Jampani}, \bibinfo{person}{Wei Wei}, \bibinfo{person}{Tingbo Hou}, \bibinfo{person}{Yael Pritch}, \bibinfo{person}{Neal Wadhwa}, \bibinfo{person}{Michael Rubinstein}, {and} \bibinfo{person}{Kfir Aberman}.} \bibinfo{year}{2023}\natexlab{b}.
\newblock \showarticletitle{Hyperdreambooth: Hypernetworks for fast personalization of text-to-image models}.
\newblock \bibinfo{journal}{\emph{arXiv preprint arXiv:2307.06949}} (\bibinfo{year}{2023}).
\newblock


\bibitem[Saharia et~al\mbox{.}(2022)]%
        {Imagen}
\bibfield{author}{\bibinfo{person}{Chitwan Saharia}, \bibinfo{person}{William Chan}, \bibinfo{person}{Saurabh Saxena}, \bibinfo{person}{Lala Li}, \bibinfo{person}{Jay Whang}, \bibinfo{person}{Emily~L Denton}, \bibinfo{person}{Kamyar Ghasemipour}, \bibinfo{person}{Raphael Gontijo~Lopes}, \bibinfo{person}{Burcu Karagol~Ayan}, \bibinfo{person}{Tim Salimans}, {et~al\mbox{.}}} \bibinfo{year}{2022}\natexlab{}.
\newblock \showarticletitle{Photorealistic text-to-image diffusion models with deep language understanding}.
\newblock \bibinfo{journal}{\emph{Advances in Neural Information Processing Systems}}  \bibinfo{volume}{35} (\bibinfo{year}{2022}), \bibinfo{pages}{36479--36494}.
\newblock


\bibitem[Sha et~al\mbox{.}(2023)]%
        {de_fake}
\bibfield{author}{\bibinfo{person}{Zeyang Sha}, \bibinfo{person}{Zheng Li}, \bibinfo{person}{Ning Yu}, {and} \bibinfo{person}{Yang Zhang}.} \bibinfo{year}{2023}\natexlab{}.
\newblock \showarticletitle{De-fake: Detection and attribution of fake images generated by text-to-image generation models}. In \bibinfo{booktitle}{\emph{Proceedings of the 2023 ACM SIGSAC Conference on Computer and Communications Security}}. \bibinfo{pages}{3418--3432}.
\newblock


\bibitem[Shi et~al\mbox{.}(2023b)]%
        {instantbooth}
\bibfield{author}{\bibinfo{person}{Jing Shi}, \bibinfo{person}{Wei Xiong}, \bibinfo{person}{Zhe Lin}, {and} \bibinfo{person}{Hyun~Joon Jung}.} \bibinfo{year}{2023}\natexlab{b}.
\newblock \showarticletitle{Instantbooth: Personalized text-to-image generation without test-time finetuning}.
\newblock \bibinfo{journal}{\emph{arXiv preprint arXiv:2304.03411}} (\bibinfo{year}{2023}).
\newblock


\bibitem[Shi et~al\mbox{.}(2023a)]%
        {mvdream}
\bibfield{author}{\bibinfo{person}{Yichun Shi}, \bibinfo{person}{Peng Wang}, \bibinfo{person}{Jianglong Ye}, \bibinfo{person}{Mai Long}, \bibinfo{person}{Kejie Li}, {and} \bibinfo{person}{Xiao Yang}.} \bibinfo{year}{2023}\natexlab{a}.
\newblock \showarticletitle{Mvdream: Multi-view diffusion for 3d generation}.
\newblock \bibinfo{journal}{\emph{arXiv preprint arXiv:2308.16512}} (\bibinfo{year}{2023}).
\newblock


\bibitem[Smith et~al\mbox{.}(2023)]%
        {continual_diffusion}
\bibfield{author}{\bibinfo{person}{James~Seale Smith}, \bibinfo{person}{Yen-Chang Hsu}, \bibinfo{person}{Lingyu Zhang}, \bibinfo{person}{Ting Hua}, \bibinfo{person}{Zsolt Kira}, \bibinfo{person}{Yilin Shen}, {and} \bibinfo{person}{Hongxia Jin}.} \bibinfo{year}{2023}\natexlab{}.
\newblock \showarticletitle{Continual diffusion: Continual customization of text-to-image diffusion with c-lora}.
\newblock \bibinfo{journal}{\emph{arXiv preprint arXiv:2304.06027}} (\bibinfo{year}{2023}).
\newblock


\bibitem[Song et~al\mbox{.}(2020)]%
        {DDIM}
\bibfield{author}{\bibinfo{person}{Jiaming Song}, \bibinfo{person}{Chenlin Meng}, {and} \bibinfo{person}{Stefano Ermon}.} \bibinfo{year}{2020}\natexlab{}.
\newblock \showarticletitle{Denoising diffusion implicit models}.
\newblock \bibinfo{journal}{\emph{arXiv preprint arXiv:2010.02502}} (\bibinfo{year}{2020}).
\newblock


\bibitem[Tewel et~al\mbox{.}(2023)]%
        {key_locked_rank}
\bibfield{author}{\bibinfo{person}{Yoad Tewel}, \bibinfo{person}{Rinon Gal}, \bibinfo{person}{Gal Chechik}, {and} \bibinfo{person}{Yuval Atzmon}.} \bibinfo{year}{2023}\natexlab{}.
\newblock \showarticletitle{Key-locked rank one editing for text-to-image personalization}. In \bibinfo{booktitle}{\emph{ACM SIGGRAPH 2023 Conference Proceedings}}. \bibinfo{pages}{1--11}.
\newblock


\bibitem[Touvron et~al\mbox{.}(2023)]%
        {llama2}
\bibfield{author}{\bibinfo{person}{Hugo Touvron}, \bibinfo{person}{Louis Martin}, \bibinfo{person}{Kevin Stone}, \bibinfo{person}{Peter Albert}, \bibinfo{person}{Amjad Almahairi}, \bibinfo{person}{Yasmine Babaei}, \bibinfo{person}{Nikolay Bashlykov}, \bibinfo{person}{Soumya Batra}, \bibinfo{person}{Prajjwal Bhargava}, \bibinfo{person}{Shruti Bhosale}, {et~al\mbox{.}}} \bibinfo{year}{2023}\natexlab{}.
\newblock \showarticletitle{Llama 2: Open foundation and fine-tuned chat models}.
\newblock \bibinfo{journal}{\emph{arXiv preprint arXiv:2307.09288}} (\bibinfo{year}{2023}).
\newblock


\bibitem[Tumanyan et~al\mbox{.}(2023)]%
        {image_to_image}
\bibfield{author}{\bibinfo{person}{Narek Tumanyan}, \bibinfo{person}{Michal Geyer}, \bibinfo{person}{Shai Bagon}, {and} \bibinfo{person}{Tali Dekel}.} \bibinfo{year}{2023}\natexlab{}.
\newblock \showarticletitle{Plug-and-play diffusion features for text-driven image-to-image translation}. In \bibinfo{booktitle}{\emph{Proceedings of the IEEE/CVF Conference on Computer Vision and Pattern Recognition}}. \bibinfo{pages}{1921--1930}.
\newblock


\bibitem[Voynov et~al\mbox{.}(2023a)]%
        {voynov2023sketch}
\bibfield{author}{\bibinfo{person}{Andrey Voynov}, \bibinfo{person}{Kfir Aberman}, {and} \bibinfo{person}{Daniel Cohen-Or}.} \bibinfo{year}{2023}\natexlab{a}.
\newblock \showarticletitle{Sketch-guided text-to-image diffusion models}. In \bibinfo{booktitle}{\emph{ACM SIGGRAPH 2023 Conference Proceedings}}. \bibinfo{pages}{1--11}.
\newblock


\bibitem[Voynov et~al\mbox{.}(2023b)]%
        {p+}
\bibfield{author}{\bibinfo{person}{Andrey Voynov}, \bibinfo{person}{Qinghao Chu}, \bibinfo{person}{Daniel Cohen-Or}, {and} \bibinfo{person}{Kfir Aberman}.} \bibinfo{year}{2023}\natexlab{b}.
\newblock \showarticletitle{$ P+ $: Extended Textual Conditioning in Text-to-Image Generation}.
\newblock \bibinfo{journal}{\emph{arXiv preprint arXiv:2303.09522}} (\bibinfo{year}{2023}).
\newblock


\bibitem[Wang et~al\mbox{.}(2023)]%
        {rodin}
\bibfield{author}{\bibinfo{person}{Tengfei Wang}, \bibinfo{person}{Bo Zhang}, \bibinfo{person}{Ting Zhang}, \bibinfo{person}{Shuyang Gu}, \bibinfo{person}{Jianmin Bao}, \bibinfo{person}{Tadas Baltrusaitis}, \bibinfo{person}{Jingjing Shen}, \bibinfo{person}{Dong Chen}, \bibinfo{person}{Fang Wen}, \bibinfo{person}{Qifeng Chen}, {et~al\mbox{.}}} \bibinfo{year}{2023}\natexlab{}.
\newblock \showarticletitle{Rodin: A generative model for sculpting 3d digital avatars using diffusion}. In \bibinfo{booktitle}{\emph{Proceedings of the IEEE/CVF Conference on Computer Vision and Pattern Recognition}}. \bibinfo{pages}{4563--4573}.
\newblock


\bibitem[Wei et~al\mbox{.}(2023)]%
        {elite}
\bibfield{author}{\bibinfo{person}{Yuxiang Wei}, \bibinfo{person}{Yabo Zhang}, \bibinfo{person}{Zhilong Ji}, \bibinfo{person}{Jinfeng Bai}, \bibinfo{person}{Lei Zhang}, {and} \bibinfo{person}{Wangmeng Zuo}.} \bibinfo{year}{2023}\natexlab{}.
\newblock \showarticletitle{Elite: Encoding visual concepts into textual embeddings for customized text-to-image generation}.
\newblock \bibinfo{journal}{\emph{arXiv preprint arXiv:2302.13848}} (\bibinfo{year}{2023}).
\newblock


\bibitem[Wu et~al\mbox{.}(2023b)]%
        {tune_a_video}
\bibfield{author}{\bibinfo{person}{Jay~Zhangjie Wu}, \bibinfo{person}{Yixiao Ge}, \bibinfo{person}{Xintao Wang}, \bibinfo{person}{Stan~Weixian Lei}, \bibinfo{person}{Yuchao Gu}, \bibinfo{person}{Yufei Shi}, \bibinfo{person}{Wynne Hsu}, \bibinfo{person}{Ying Shan}, \bibinfo{person}{Xiaohu Qie}, {and} \bibinfo{person}{Mike~Zheng Shou}.} \bibinfo{year}{2023}\natexlab{b}.
\newblock \showarticletitle{Tune-a-video: One-shot tuning of image diffusion models for text-to-video generation}. In \bibinfo{booktitle}{\emph{Proceedings of the IEEE/CVF International Conference on Computer Vision}}. \bibinfo{pages}{7623--7633}.
\newblock


\bibitem[Wu et~al\mbox{.}(2023a)]%
        {multimodal_llm}
\bibfield{author}{\bibinfo{person}{Shengqiong Wu}, \bibinfo{person}{Hao Fei}, \bibinfo{person}{Leigang Qu}, \bibinfo{person}{Wei Ji}, {and} \bibinfo{person}{Tat-Seng Chua}.} \bibinfo{year}{2023}\natexlab{a}.
\newblock \showarticletitle{Next-gpt: Any-to-any multimodal llm}.
\newblock \bibinfo{journal}{\emph{arXiv preprint arXiv:2309.05519}} (\bibinfo{year}{2023}).
\newblock


\bibitem[Xu et~al\mbox{.}(2023a)]%
        {dream3d}
\bibfield{author}{\bibinfo{person}{Jiale Xu}, \bibinfo{person}{Xintao Wang}, \bibinfo{person}{Weihao Cheng}, \bibinfo{person}{Yan-Pei Cao}, \bibinfo{person}{Ying Shan}, \bibinfo{person}{Xiaohu Qie}, {and} \bibinfo{person}{Shenghua Gao}.} \bibinfo{year}{2023}\natexlab{a}.
\newblock \showarticletitle{Dream3d: Zero-shot text-to-3d synthesis using 3d shape prior and text-to-image diffusion models}. In \bibinfo{booktitle}{\emph{Proceedings of the IEEE/CVF Conference on Computer Vision and Pattern Recognition}}. \bibinfo{pages}{20908--20918}.
\newblock


\bibitem[Xu et~al\mbox{.}(2018)]%
        {AttnGAN}
\bibfield{author}{\bibinfo{person}{Tao Xu}, \bibinfo{person}{Pengchuan Zhang}, \bibinfo{person}{Qiuyuan Huang}, \bibinfo{person}{Han Zhang}, \bibinfo{person}{Zhe Gan}, \bibinfo{person}{Xiaolei Huang}, {and} \bibinfo{person}{Xiaodong He}.} \bibinfo{year}{2018}\natexlab{}.
\newblock \showarticletitle{Attngan: Fine-grained text to image generation with attentional generative adversarial networks}. In \bibinfo{booktitle}{\emph{Proceedings of the IEEE conference on computer vision and pattern recognition}}. \bibinfo{pages}{1316--1324}.
\newblock


\bibitem[Xu et~al\mbox{.}(2023b)]%
        {versatile_diffusion}
\bibfield{author}{\bibinfo{person}{Xingqian Xu}, \bibinfo{person}{Zhangyang Wang}, \bibinfo{person}{Gong Zhang}, \bibinfo{person}{Kai Wang}, {and} \bibinfo{person}{Humphrey Shi}.} \bibinfo{year}{2023}\natexlab{b}.
\newblock \showarticletitle{Versatile diffusion: Text, images and variations all in one diffusion model}. In \bibinfo{booktitle}{\emph{Proceedings of the IEEE/CVF International Conference on Computer Vision}}. \bibinfo{pages}{7754--7765}.
\newblock


\bibitem[Xue et~al\mbox{.}(2024)]%
        {raphael}
\bibfield{author}{\bibinfo{person}{Zeyue Xue}, \bibinfo{person}{Guanglu Song}, \bibinfo{person}{Qiushan Guo}, \bibinfo{person}{Boxiao Liu}, \bibinfo{person}{Zhuofan Zong}, \bibinfo{person}{Yu Liu}, {and} \bibinfo{person}{Ping Luo}.} \bibinfo{year}{2024}\natexlab{}.
\newblock \showarticletitle{Raphael: Text-to-image generation via large mixture of diffusion paths}.
\newblock \bibinfo{journal}{\emph{Advances in Neural Information Processing Systems}}  \bibinfo{volume}{36} (\bibinfo{year}{2024}).
\newblock


\bibitem[Yang et~al\mbox{.}(2023a)]%
        {paint_by_example}
\bibfield{author}{\bibinfo{person}{Binxin Yang}, \bibinfo{person}{Shuyang Gu}, \bibinfo{person}{Bo Zhang}, \bibinfo{person}{Ting Zhang}, \bibinfo{person}{Xuejin Chen}, \bibinfo{person}{Xiaoyan Sun}, \bibinfo{person}{Dong Chen}, {and} \bibinfo{person}{Fang Wen}.} \bibinfo{year}{2023}\natexlab{a}.
\newblock \showarticletitle{Paint by example: Exemplar-based image editing with diffusion models}. In \bibinfo{booktitle}{\emph{Proceedings of the IEEE/CVF Conference on Computer Vision and Pattern Recognition}}. \bibinfo{pages}{18381--18391}.
\newblock


\bibitem[Yang et~al\mbox{.}(2023b)]%
        {diffusion_survey}
\bibfield{author}{\bibinfo{person}{Ling Yang}, \bibinfo{person}{Zhilong Zhang}, \bibinfo{person}{Yang Song}, \bibinfo{person}{Shenda Hong}, \bibinfo{person}{Runsheng Xu}, \bibinfo{person}{Yue Zhao}, \bibinfo{person}{Wentao Zhang}, \bibinfo{person}{Bin Cui}, {and} \bibinfo{person}{Ming-Hsuan Yang}.} \bibinfo{year}{2023}\natexlab{b}.
\newblock \showarticletitle{Diffusion models: A comprehensive survey of methods and applications}.
\newblock \bibinfo{journal}{\emph{Comput. Surveys}} \bibinfo{volume}{56}, \bibinfo{number}{4} (\bibinfo{year}{2023}), \bibinfo{pages}{1--39}.
\newblock


\bibitem[Yao et~al\mbox{.}(2024)]%
        {llm_security_survey}
\bibfield{author}{\bibinfo{person}{Yifan Yao}, \bibinfo{person}{Jinhao Duan}, \bibinfo{person}{Kaidi Xu}, \bibinfo{person}{Yuanfang Cai}, \bibinfo{person}{Zhibo Sun}, {and} \bibinfo{person}{Yue Zhang}.} \bibinfo{year}{2024}\natexlab{}.
\newblock \showarticletitle{A survey on large language model (llm) security and privacy: The good, the bad, and the ugly}.
\newblock \bibinfo{journal}{\emph{High-Confidence Computing}} (\bibinfo{year}{2024}), \bibinfo{pages}{100211}.
\newblock


\bibitem[Zhang et~al\mbox{.}(2023b)]%
        {text-to-image-survey}
\bibfield{author}{\bibinfo{person}{Chenshuang Zhang}, \bibinfo{person}{Chaoning Zhang}, \bibinfo{person}{Mengchun Zhang}, {and} \bibinfo{person}{In~So Kweon}.} \bibinfo{year}{2023}\natexlab{b}.
\newblock \showarticletitle{Text-to-image diffusion model in generative ai: A survey}.
\newblock \bibinfo{journal}{\emph{arXiv preprint arXiv:2303.07909}} (\bibinfo{year}{2023}).
\newblock


\bibitem[Zhang et~al\mbox{.}(2017)]%
        {StackGAN}
\bibfield{author}{\bibinfo{person}{Han Zhang}, \bibinfo{person}{Tao Xu}, \bibinfo{person}{Hongsheng Li}, \bibinfo{person}{Shaoting Zhang}, \bibinfo{person}{Xiaogang Wang}, \bibinfo{person}{Xiaolei Huang}, {and} \bibinfo{person}{Dimitris~N Metaxas}.} \bibinfo{year}{2017}\natexlab{}.
\newblock \showarticletitle{Stackgan: Text to photo-realistic image synthesis with stacked generative adversarial networks}. In \bibinfo{booktitle}{\emph{Proceedings of the IEEE international conference on computer vision}}. \bibinfo{pages}{5907--5915}.
\newblock


\bibitem[Zhang et~al\mbox{.}(2023a)]%
        {controlnet}
\bibfield{author}{\bibinfo{person}{Lvmin Zhang}, \bibinfo{person}{Anyi Rao}, {and} \bibinfo{person}{Maneesh Agrawala}.} \bibinfo{year}{2023}\natexlab{a}.
\newblock \showarticletitle{Adding conditional control to text-to-image diffusion models}. In \bibinfo{booktitle}{\emph{Proceedings of the IEEE/CVF International Conference on Computer Vision}}. \bibinfo{pages}{3836--3847}.
\newblock


\bibitem[Zhao et~al\mbox{.}(2024)]%
        {uni-controlnet}
\bibfield{author}{\bibinfo{person}{Shihao Zhao}, \bibinfo{person}{Dongdong Chen}, \bibinfo{person}{Yen-Chun Chen}, \bibinfo{person}{Jianmin Bao}, \bibinfo{person}{Shaozhe Hao}, \bibinfo{person}{Lu Yuan}, {and} \bibinfo{person}{Kwan-Yee~K Wong}.} \bibinfo{year}{2024}\natexlab{}.
\newblock \showarticletitle{Uni-controlnet: All-in-one control to text-to-image diffusion models}.
\newblock \bibinfo{journal}{\emph{Advances in Neural Information Processing Systems}}  \bibinfo{volume}{36} (\bibinfo{year}{2024}).
\newblock


\bibitem[Zheng et~al\mbox{.}(2024)]%
        {judging_llm}
\bibfield{author}{\bibinfo{person}{Lianmin Zheng}, \bibinfo{person}{Wei-Lin Chiang}, \bibinfo{person}{Ying Sheng}, \bibinfo{person}{Siyuan Zhuang}, \bibinfo{person}{Zhanghao Wu}, \bibinfo{person}{Yonghao Zhuang}, \bibinfo{person}{Zi Lin}, \bibinfo{person}{Zhuohan Li}, \bibinfo{person}{Dacheng Li}, \bibinfo{person}{Eric Xing}, {et~al\mbox{.}}} \bibinfo{year}{2024}\natexlab{}.
\newblock \showarticletitle{Judging llm-as-a-judge with mt-bench and chatbot arena}.
\newblock \bibinfo{journal}{\emph{Advances in Neural Information Processing Systems}}  \bibinfo{volume}{36} (\bibinfo{year}{2024}).
\newblock


\bibitem[Zhuang et~al\mbox{.}(2023)]%
        {dreameditor}
\bibfield{author}{\bibinfo{person}{Jingyu Zhuang}, \bibinfo{person}{Chen Wang}, \bibinfo{person}{Liang Lin}, \bibinfo{person}{Lingjie Liu}, {and} \bibinfo{person}{Guanbin Li}.} \bibinfo{year}{2023}\natexlab{}.
\newblock \showarticletitle{Dreameditor: Text-driven 3d scene editing with neural fields}. In \bibinfo{booktitle}{\emph{SIGGRAPH Asia 2023 Conference Papers}}. \bibinfo{pages}{1--10}.
\newblock


\end{thebibliography}

\appendix

\section{Complete Prompt Template for ChatGPT}

Here is the complete prompt template  designed for ChatGPT to conduct semantic analysis:

\begin{quote}

  I will give you some examples. 

  Given a sentence "arafed dog sitting on the beach with its tongue out", the foreground is "arafed dog with its tongue out", the background is "beach" and the action is "sitting". 
  
  Given a sentence "there is a cat and a dog that are playing together", the foreground is "a cat and a dog", the background is "None" and the action is "playing". 
  
  Given a sentence "there are two cats that are playing with each other", the foreground is "two cats", the background is "None" and the action is "playing with each other". 
  
  Given a sentence "there is a cat on the bench", the foreground is "a cat", the background is "bench" and the action is "None". 
  
  Given a sentence "there is a white sandy beach with cat", the foreground is "a white sandy beach with cat", the background is "None" and the action is "None". 
  
  Now imitate these, I will give you a sentence "xxx" here, and you need to give me the foreground, background and action.

  \textit{(xxx here is the text description $T_c$)}
\end{quote}


Initially, we provide ChatGPT with five examples to enhance its understanding of semantic analysis. Subsequently, we present it with the text description $T_c$ to encourage it to emulate these examples. Given that $T_c$ is generated from BLIP and may have diverse formats, we expose ChatGPT to instances where neither the background nor action is present. For such cases, we instruct ChatGPT to respond with ``None''. The decision not to impose a strict response format on ChatGPT is intentional, as our observations suggest that enforcing strict formatting does not ensure consistency and can potentially compromise the accuracy of the results.

\section{Application Based on Custom Diffusion in the Finetuning Process}


In this section, we demonstrate the application of our paradigm based on Custom Diffusion \cite{custom-diffusion} in the finetuning process. The application of Custom Diffusion shares similarities with that of Dreambooth \cite{dreambooth}. Both approaches utilize the finetuning process to tailor concepts while retaining detailed prior knowledge. Our focus is on highlighting the distinctions between them.
Dreambooth uses fixed text embeddings of the rare token $[V]$ in its methods, while Custom Diffusion updates the text embeddings of the rare token $[V]$ during the finetuning process. Additionally, in the Custom Diffusion-based application, we finetune only the parameters of the cross-attention block and text embeddings of the rare token, instead of all parameters of the diffusion model. Notably, the original Custom Diffusion employs a retrieval process to fetch pairs of text prompts and images similar to the text prompt $P_t$. In our experiments, we replace this with a generation process. The Custom Diffusion version incorporates additional techniques to enhance the finetuning process's performance. For instance, it involves randomly resizing images and appending prompts like ``very small'', ``far away'', or ``zoomed in'', ``close up'' to the text prompt based on the resize ratio. For further details, refer to the original paper \cite{custom-diffusion}.

\section{Accuracy of Main Object's Extracted Description}

\begin{figure}[t]
  \centering
  \includegraphics[width=\linewidth]{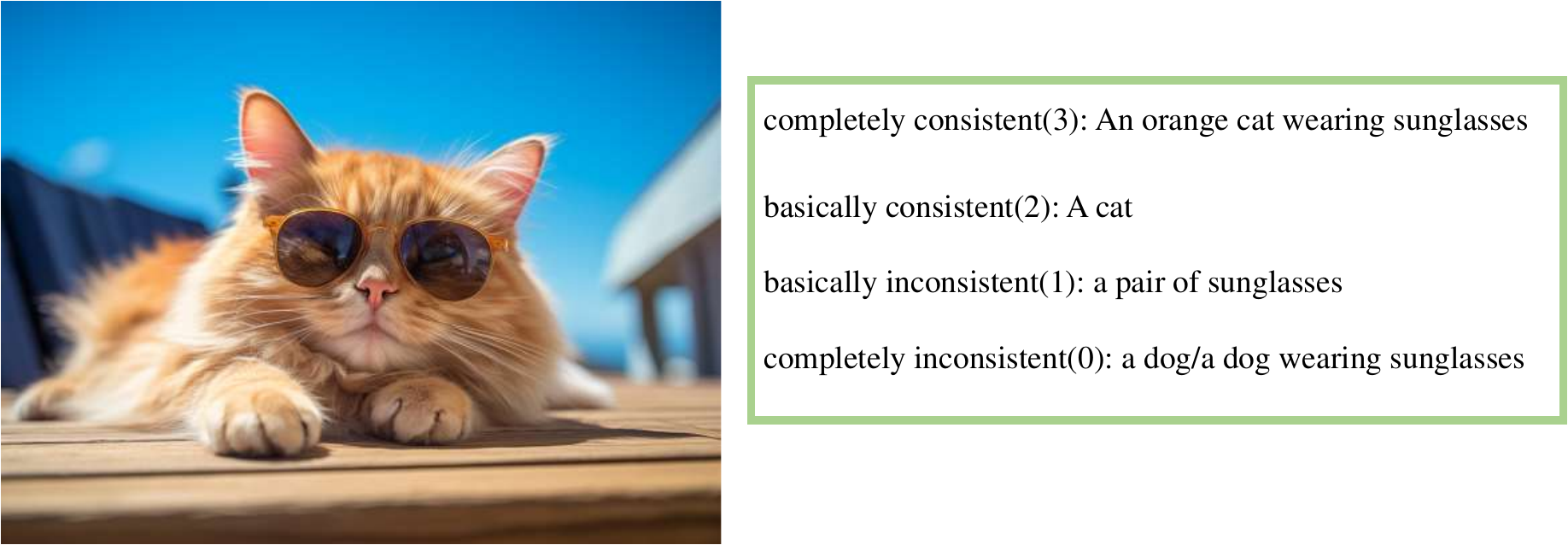}
  \caption{\textbf{Sample guide for evaluation of Main Object Description Extraction.} Participants are requested to classify each pair of an image and its extracted description into one of four categories: completely consistent, basically consistent, basically inconsistent, and completely inconsistent. Scores range from 3 to 0, corresponding to these categories in descending order of consistency.}
  \label{extraction_evaluation}
\end{figure}
\begin{figure}[t]
  \centering
  \includegraphics[width=0.8\linewidth]{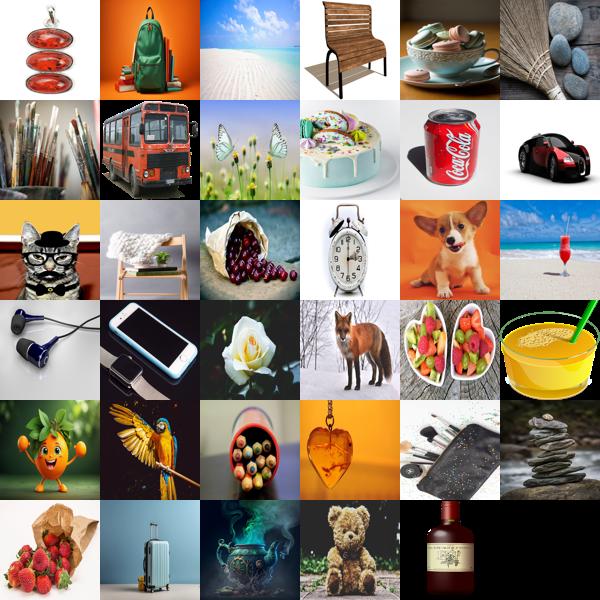}
  \caption{\textbf{All 35 Images for Evaluation of Main Object Description Extraction.} Illustrated here are all 35 images used for the evaluation of the main object description extraction.}
  \label{all_images}
\end{figure}

In this section, we assess the accuracy of the main object descriptions extracted in the initial phase of our paradigm. Utilizing 35 images featuring distinct main objects, we extract descriptions for each main object in the images. We present these image-description pairs to 50 individuals and solicit their input on the appropriateness of the extracted descriptions. Specifically, we request them to categorize each pair into one of four classes: completely consistent, basically consistent, basically inconsistent, and completely inconsistent. A ``completely consistent'' classification indicates that the extracted description provides an accurate and detailed account of the main object in the image. ``Basically consistent'' implies that the extracted description is generally accurate but lacks detailed information. ``Basically inconsistent'' signifies that the extracted description describes objects in the image, but not the main object. Finally, ``completely inconsistent'' suggests that the extracted description inaccurately describes an object not relevant to the main object in the image. 
In \cref{extraction_evaluation}, we display a sample guide given to the participants for their evaluation. 
All 35 images for the evaluation are shown in \cref{all_images}.

The results are presented in \cref{table_extraction_evaluation}. In this assessment, the majority of respondents indicated that the extracted descriptions of the main objects are categorized as ``completely consistent''. Remarkably, there was a consensus among all participants that the descriptions fell into either the ``completely consistent'' or ``basically consistent''. These findings affirm the reasonability of the extracted descriptions of the main objects. Therefore, utilizing these descriptions in the second phase of our paradigm, as well as in the text-alignment evaluation, is acknowledged rational.

\begin{table}[t]
  \centering
  \begin{tabular}{c|c}
    \toprule
    Category & Percentage \\
    \midrule
    Completely consistent & 85.40\% \\
    Basically consistent & 14.60\% \\
    Basically inconsistent & 00.00\% \\
    Completely inconsistent & 00.00\% \\
    \bottomrule
  \end{tabular}
  \caption{\textbf{Human Evaluation of Main Object Description Extraction.} Illustrated here are the percentages for the four categorizations chosen by participants for the extracted descriptions of the main objects. The majority of participants found the descriptions to be completely consistent, while a smaller fraction deemed them basically consistent. Notably, no participants categorized the extracted descriptions as either basically inconsistent or completely inconsistent.}
  \label{table_extraction_evaluation}
\end{table}

\begin{figure}[ht]
  \centering
  \includegraphics[width=\linewidth]{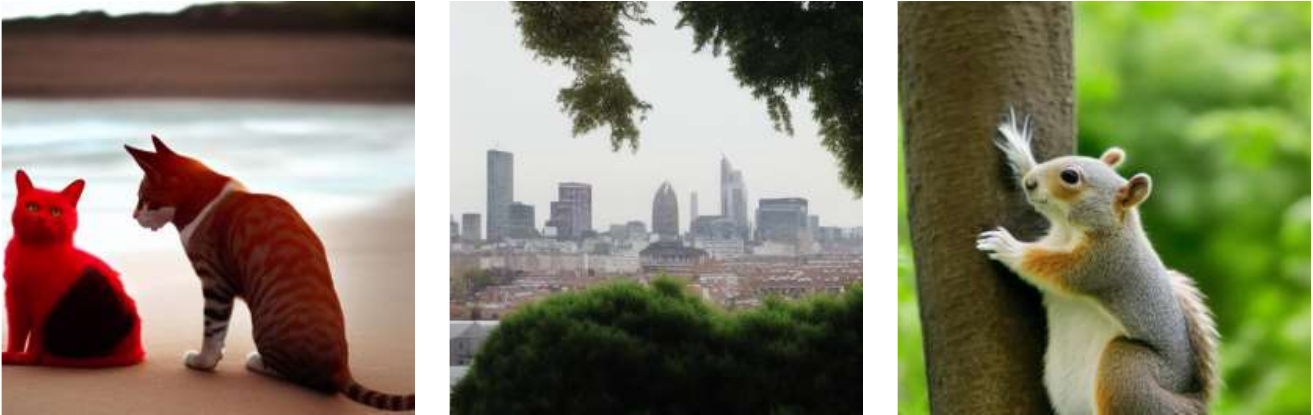}
  \caption{\textbf{Suboptimal Examples from Stable Diffusion Version 1.4.} Displayed are suboptimal outputs generated by Stable Diffusion version 1.4. Three images demonstrate the limitations of the model, each corresponding to distinct prompts from left to right: ``A photo of a cat with red color and a dog sitting on the beach'', ``A photo of a clock and a tree with a city in the background'', ``A photo of a teddy bear and a squirrel in the jungle''. The first image inaccurately represents the concept of a dog, depicting two cats instead. The second image misses the concept of a clock, and the third inaccurately captures the concept of a teddy bear.}
  \label{limitation_of_SD}
\end{figure}

\begin{figure}[ht]
  \centering
  \includegraphics[width=\linewidth]{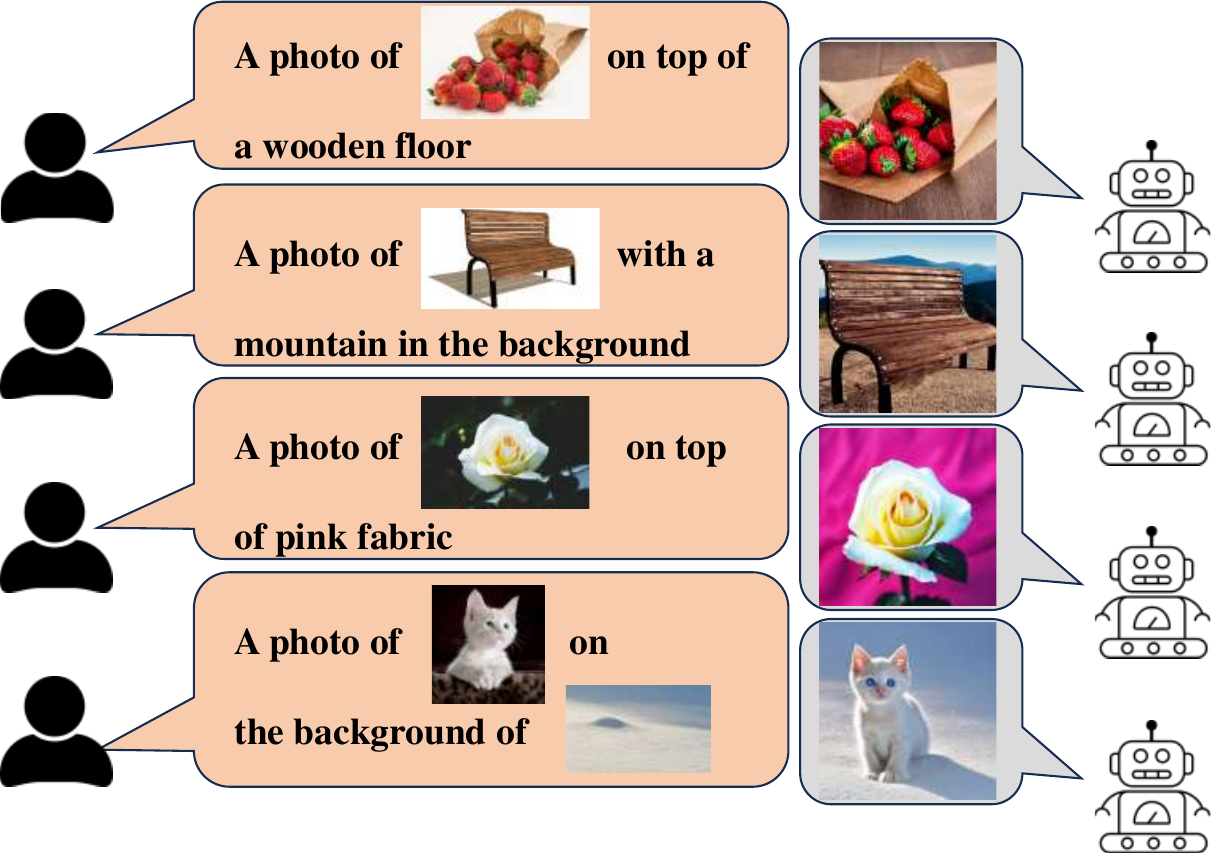}
  \caption{\textbf{Additional Results.} Additional results of customized generation with multi-modal prompts.}
  \label{more_results}
\end{figure}

\section{Limitations of Current Stable Diffusions}


Current stable diffusion models demonstrate impressive performance in simple text-to-image generation tasks, yet they exhibit limitations when faced with more complex text prompts. Taking Stable Diffusion version 1.4 \cite{SD} as an example, we showcase some suboptimal results generated from this model in \cref{limitation_of_SD}. It becomes evident that when confronted with intricate text prompts containing multiple objects, Stable Diffusion version 1.4 may struggle to accurately generate the concepts of some objects. It may either produce incorrect objects or overlook certain elements. Similar challenges arise when attempting customized text-to-image generation through finetuning stable diffusion. This is why, in our experiments, we exclusively explore the scenario involving a single image within the multi-modal prompt. SDXL \cite{SDXL} represents a more advanced diffusion model that holds the promise of improved performance in generating images from complex text prompts. Future work will investigate the utilization of SDXL to assess the performance of our paradigm in scenarios involving multiple images.

\section{Additional Results of Customized Generation with Multi-Modal Prompts}


In the main paper, we provided a subset of the results from customized generation with multi-modal prompts, limited by space constraints. Here, we present additional results of customized generation with multi-modal prompts in \cref{more_results}. These visual representations reaffirm our paradigm's ability to generate images that harmonize with both the text prompts and the images within the multi-modal prompts. Once again, these findings collectively emphasize the effectiveness of our paradigm in customized generation with multi-modal prompts.

\end{document}